\documentclass{article}

\usepackage[preprint]{neurips_2026}

\usepackage[utf8]{inputenc}
\usepackage[T1]{fontenc}
\usepackage{hyperref}
\usepackage{url}
\usepackage{booktabs}
\usepackage{amsfonts}
\usepackage{amsmath}
\usepackage{amssymb}
\usepackage{nicefrac}
\usepackage{microtype}
\usepackage{xcolor}
\usepackage{graphicx}
\usepackage{placeins}
\usepackage[most]{tcolorbox}
\usepackage[ruled,vlined]{algorithm2e}
\setcitestyle{numbers,square,comma,sort&compress}
\hypersetup{
  colorlinks=true,
  linkcolor=blue!45!black,
  citecolor=blue!45!black,
  urlcolor=blue!45!black,
  pdftitle={FaithSieve: Fine-Grained Evaluation of Math Proofs with Faithful Formal Evidence},
  pdfauthor={Ziyu Wang, Qiming Dai, Yishan Wu, Zaiwen Wen}
}

\newcommand{\arxivpublic}{}

\newtcblisting{PromptBox}{
  listing only,
  breakable,
  colback=gray!5,
  colframe=black!50,
  boxrule=0.4pt,
  arc=1mm,
  left=2mm,
  right=2mm,
  top=2mm,
  bottom=2mm,
  listing options={
    basicstyle=\ttfamily\footnotesize,
    columns=fullflexible,
    keepspaces=true,
    showstringspaces=false,
    breaklines=true
  }
}

\title{FaithSieve: Fine-Grained Evaluation of Math Proofs with Faithful Formal Evidence}

\author{%
  Ziyu Wang\\
  Academy for Advanced\\
  Interdisciplinary Studies\\
  Peking University\\
  \texttt{wangziyu-edu@stu.pku.edu.cn}\\
  \And
  Qiming Dai\\
  School of Mathematical Sciences\\
  Peking University\\
  \texttt{qmdai25@stu.pku.edu.cn}\\
  \AND
  Yishan Wu\\
  Theory Lab, 2012 Labs\\
  Huawei Technologies Co., Ltd.\\
  \texttt{wuyishan1@huawei.com}\\
  \And
  Zaiwen Wen\\
  Beijing International Center for\\
  Mathematical Research\\
  Peking University\\
  \texttt{wenzw@pku.edu.cn}\\
}

\begin{document}

\maketitle
\addtocontents{toc}{\protect\setcounter{tocdepth}{-1}}

\begin{abstract}
Large language models can now generate complex, multi-step mathematical proofs, but reliably determining their correctness and localizing early logical errors remains a critical challenge. Existing evaluation approaches largely depend on model-based natural-language judgments, which often overlook local reasoning gaps. While formal theorem provers like Lean offer a path to rigorous verification, using them to evaluate informal text requires solving locality and semantic mismatches: a prover might bypass a local flaw by proving an overly broad target, or validate an auto-formalized statement that drifts from the original mathematical intent. To address this, we introduce FaithSieve, a Lean-assisted framework for fine-grained evaluation of natural-language mathematical proofs. FaithSieve decomposes coarse proof steps into local reasoning units, extracts typed proof obligations, and verifies them through a formal evaluation agent. Formal validation is gated by semantic alignment scoring, so Lean evidence is incorporated only when the formal statement faithfully preserves the context, objects, and logical form of the original claim. We construct two expert-verified datasets, ProofLoc-Olympiad and ProofLoc-University, to benchmark first-error localization. On the 350-problem Olympiad dataset, FaithSieve using a GPT-5.4 backbone achieves 81.43\% exact first-error accuracy, outperforming the direct-judging baseline of 72.29\%. Furthermore, on the 200-problem ProofLoc-University benchmark spanning six advanced domains, FaithSieve reaches 84.5\% exact accuracy, compared to 75.0\% for the direct judge. Our work demonstrates that decomposing proofs into fine-grained units and grounding them with faithful formal evidence significantly improves reliable evaluation of natural-language reasoning.

\end{abstract}

\section{Introduction}

Large language models can now generate multi-step mathematical proofs, but evaluating them and locating the first logical error remains difficult. Such proofs may omit conditions, skip derivations, use implicit assumptions, or continue from early mistakes to plausible-looking conclusions. Reliable evaluation must therefore check not only final answers but also local reasoning steps.

Existing natural-language verification methods have made substantial progress. Self-consistency, process supervision, process reward models, and recent work on error localization, ProcessBench, Hard2Verify, and ProofGrader / ProofBench provide useful baselines for step-level evaluation \citep{wang2023selfconsistency,lightman2023verify,tyen2024llms,zheng2025processbench,pandit2025hard2verify,ma2025proofgrader}. However, their evidence mainly comes from model-based scoring or classification over natural-language reasoning. In parallel, formal reasoning systems have also advanced rapidly. Lean, mathlib, and systems such as DeepSeek-Prover-V2, Seed-Prover, AxiomProver, Aria, and M2F show that many mathematical claims can now be checked with formal evidence \citep{mathlib2020lean,ren2025deepseekproverv2,chen2025seedprover,seed2025seedprover15,axiom2026axle,wang2025aria,wang2026m2f}.

Using Lean as a direct judge for informal proofs still faces a central obstacle. Lean only verifies the formal statement submitted to it; even if that statement is provable, it need not faithfully express the proof step being checked. Autoformalization may omit assumptions, add conditions, reverse an implication, or rewrite a contested local claim into a different statement that is provable but not equivalent. Formal evidence is therefore useful for error localization only when the validation target remains local and semantically faithful to the original claim.

\begin{figure}[!b]
\centering
\includegraphics[width=0.86\linewidth]{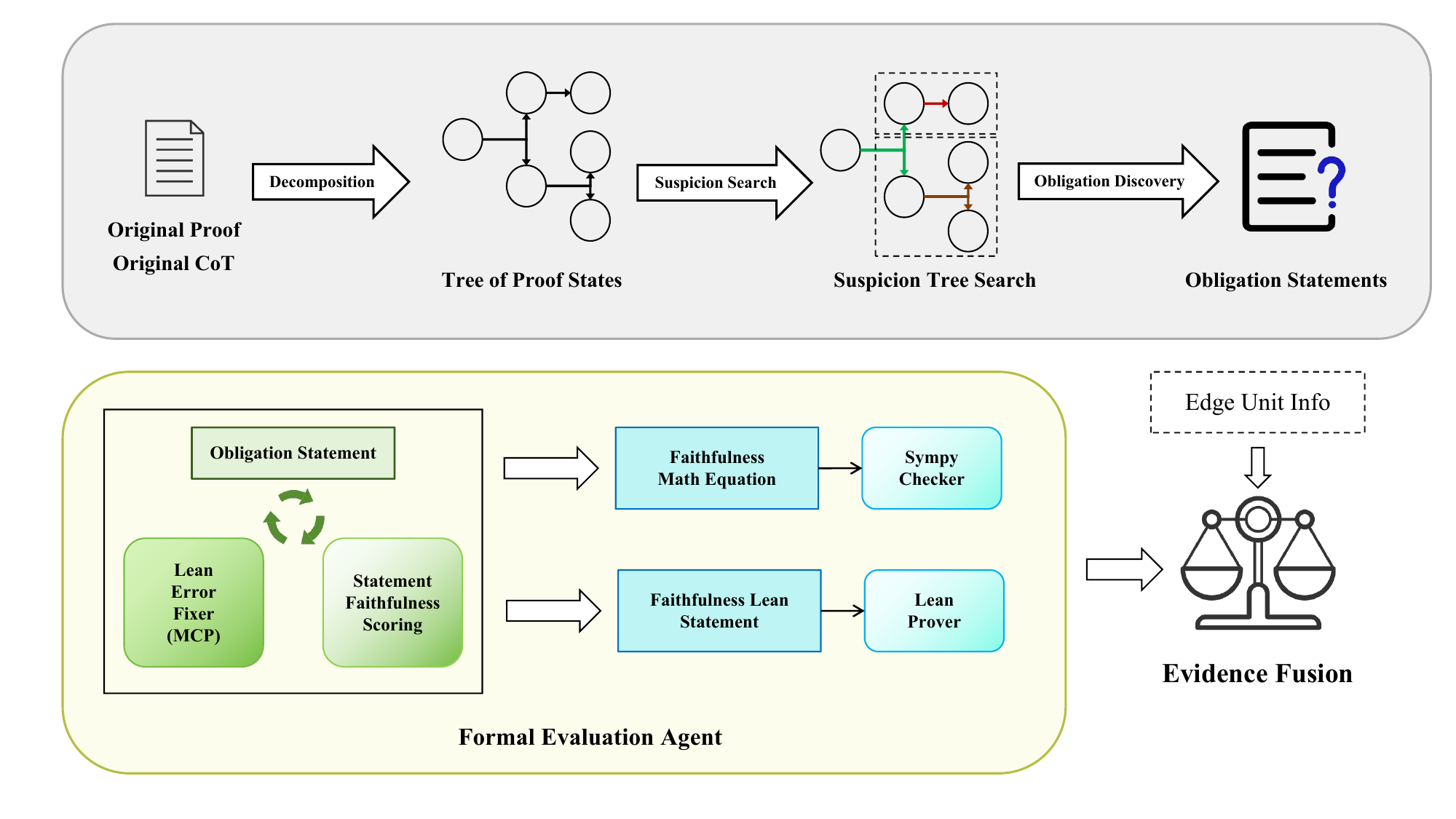}
\caption{Overview of the FaithSieve workflow for Lean-assisted first-error localization.}
\label{fig:workflow}
\end{figure}

We propose FaithSieve, a Lean-assisted framework for first-error localization in natural-language mathematical proofs. FaithSieve decomposes coarse proof steps into local \texttt{EdgeUnits}, extracts typed proof obligations from suspicious transitions, and combines lightweight checking, Lean validation, and semantic alignment scoring to produce local evidence. Lean is not used as a direct judge of the whole proof; instead, it serves as a source of structured evidence gated by semantic alignment. The overall workflow has three stages: localization, verification, and synthesis. The localization stage constructs \texttt{EdgeUnits} and selects local transitions for priority inspection. The verification stage turns these transitions into proof obligations and gathers evidence through both formal and lightweight non-formal checks. The synthesis stage aggregates local evidence back to the original step level and predicts the first erroneous step. Figure~\ref{fig:workflow} illustrates this pipeline.

We construct two expert-verified datasets: \textsc{ProofLoc-Olympiad}, with 350 algebra and number theory problems, and \textsc{ProofLoc-University}, with 200 advanced mathematics problems. On \textsc{ProofLoc-Olympiad}, FaithSieve achieves 81.43\% exact first-error accuracy, compared with 72.29\% for a GPT-5.4 direct-judging baseline. On \textsc{ProofLoc-University}, FaithSieve also obtains clear gains over direct judging. Our contributions are as follows.\\

1. \textbf{Lean-assisted proof error localization framework.} We identify locality and semantic faithfulness as the central challenges in applying Lean validation to natural-language proof evaluation, and propose FaithSieve as a Lean-assisted framework for first-error localization.\\
2. \textbf{Proof-state-tree decomposition and typed obligations.} We introduce a proof-state-tree representation that decomposes coarse proof steps into local state transitions, and extract typed obligations that specify the reasoning responsibility of each transition.\\
3. \textbf{Semantic alignment scoring.} We propose a semantic alignment score to measure whether a Lean statement preserves the intended informal obligation, so formal validation is used only when it avoids drifted or weakened targets.\\
4. \textbf{Expert-verified evaluation datasets.} We construct two expert-verified benchmarks covering olympiad-style and university-level mathematics, and show that faithful local formal evidence improves over direct judging baselines.

\section{Task Formulation and Challenges}
\label{sec:task}

\subsection{Problem Formulation}
\label{sec:problem-formulation}

When large language models generate natural-language mathematical proofs, a central challenge is not only judging overall correctness but also locating the first erroneous step. BIG-Bench Mistake first formulated first-mistake localization as an explicit evaluation problem, and later work extended it to mathematical process verification and open-ended solutions \citep{tyen2024llms,zheng2025processbench,pandit2025hard2verify}. Process supervision and PRMs provide further background for step-level correctness modeling \citep{lightman2023verify}. We adopt this formulation, but ask whether formal verification can provide checkable structured evidence for first-error localization.

Concretely, the input is a mathematical problem \(p\) and a natural-language proof \(\pi=(s_1,\ldots,s_n)\) with \(n\) steps. The system must output either \texttt{correct} or the index \(k\) of the first erroneous step, where \(k\) refers to a benchmark step. We define the first erroneous step as the earliest step that makes the proof mathematically unreliable; later errors may exist, but they are not the target.

For evaluation, we use the same label interface as prior first-error localization work \citep{tyen2024llms,zheng2025processbench} and report binary accuracy and exact accuracy. Let \(y,\hat y\in\{\texttt{correct}\}\cup\{1,\ldots,n\}\) denote the gold label and system prediction. Binary accuracy measures whether the system correctly distinguishes valid proofs from invalid ones, while exact accuracy requires an exact match to the gold label: \texttt{correct} for valid proofs, and the correct first-error step for invalid proofs.

\subsection{Lean-Based Verification: Promise and Challenges}

To go beyond purely natural-language judges, a natural direction is to use proof assistants such as Lean to provide checkable artifacts for local mathematical claims. Lean validation can turn part of the judgment from a model score into a verifiable proof result or failure signal. For algebraic rewriting, logical entailment, goal reduction, witness verification, and case coverage, formal validation may provide stronger evidence than natural-language scoring. However, this evidence is not unconditionally reliable: formal validation must target the right local object, and that object must faithfully express the original natural-language claim.

\textbf{Locality mismatch.} The first difficulty is ensuring that formal validation targets the same local inference as the natural-language proof. A proof usually consists of local reasoning transitions rather than a single theorem. If a proof claims to derive \(b\) from \(a\), and then \(c\) from \(b\), proving only \(a \to c\) does not establish that the two local steps \(a \to b\) and \(b \to c\) are correct. A strong prover may directly prove a stronger or coarser target and ignore a local error in the original proof. Formal validation therefore requires decomposition and localization: coarse steps must be broken into local reasoning units, and the system must choose which local claims to validate.

\textbf{Semantic mismatch.} The second difficulty is ensuring that the Lean statement faithfully represents the original obligation. Even after localization, the system must translate the informal claim into a formal proposition. Autoformalization may omit assumptions, add conditions, reverse implications, alter quantifiers, weaken or strengthen the conclusion, or replace the claim with a different but provable statement. Thus, compilation or proof success alone is insufficient evidence; the key question is whether the Lean statement faithfully corresponds to the original claim.

These two mismatches lead to two requirements. First, formal validation should operate on local reasoning units, so the system must decompose proofs appropriately. Second, the Lean statement being validated must be semantically aligned with the original natural-language obligation. Accordingly, we use Lean not as an end-to-end theorem prover, but as a conditional evidence generator: a formal result is strong evidence only when both locality and faithfulness are satisfied.

\section{Localizing Verifiable Reasoning Units}
\label{sec:localizing}

\subsection{Recursive Decomposition into a Tree of Proof States}

For a traditional chain-of-thought proof, a COT step is suitable as an annotation and output unit, but it is often not the smallest reasoning unit, because a single step frequently contains multiple operations simultaneously: substitution, definition expansion, introduction of new facts, goal switching, or branch pruning. If such coarse-grained steps are passed directly to a Lean pipeline, we encounter the locality mismatch discussed in Section~\ref{sec:task}. Therefore, we treat a COT step as an annotation unit, but not directly as a logical verification unit. To obtain finer-grained proofs, we first ask the language model to pre-decompose the original COT steps, removing redundant and repetitive statements without modifying the proof logic, and obtaining a detailed COT in a semantically faithful manner. Here we emphasize ``faithful'' rather than ``patching'': conclusions, variable scopes, case-discussion wording, concrete witnesses, and numerical calculations in the original proof must all be preserved as-is.

A Lean proof state records the available assumptions and current goal at a given moment, and a tactic transforms one proof state into another. Building on the fine-grained substeps obtained from preprocessing, we borrow the state-transition perspective of Lean proof states and ask the language model to write out all proof states and state transitions in the proof process step by step according to the natural-language steps, constructing a tree of proof states \(\mathcal{T}=(V,E)\).

\begin{figure}[t]
\centering
\includegraphics[width=\linewidth]{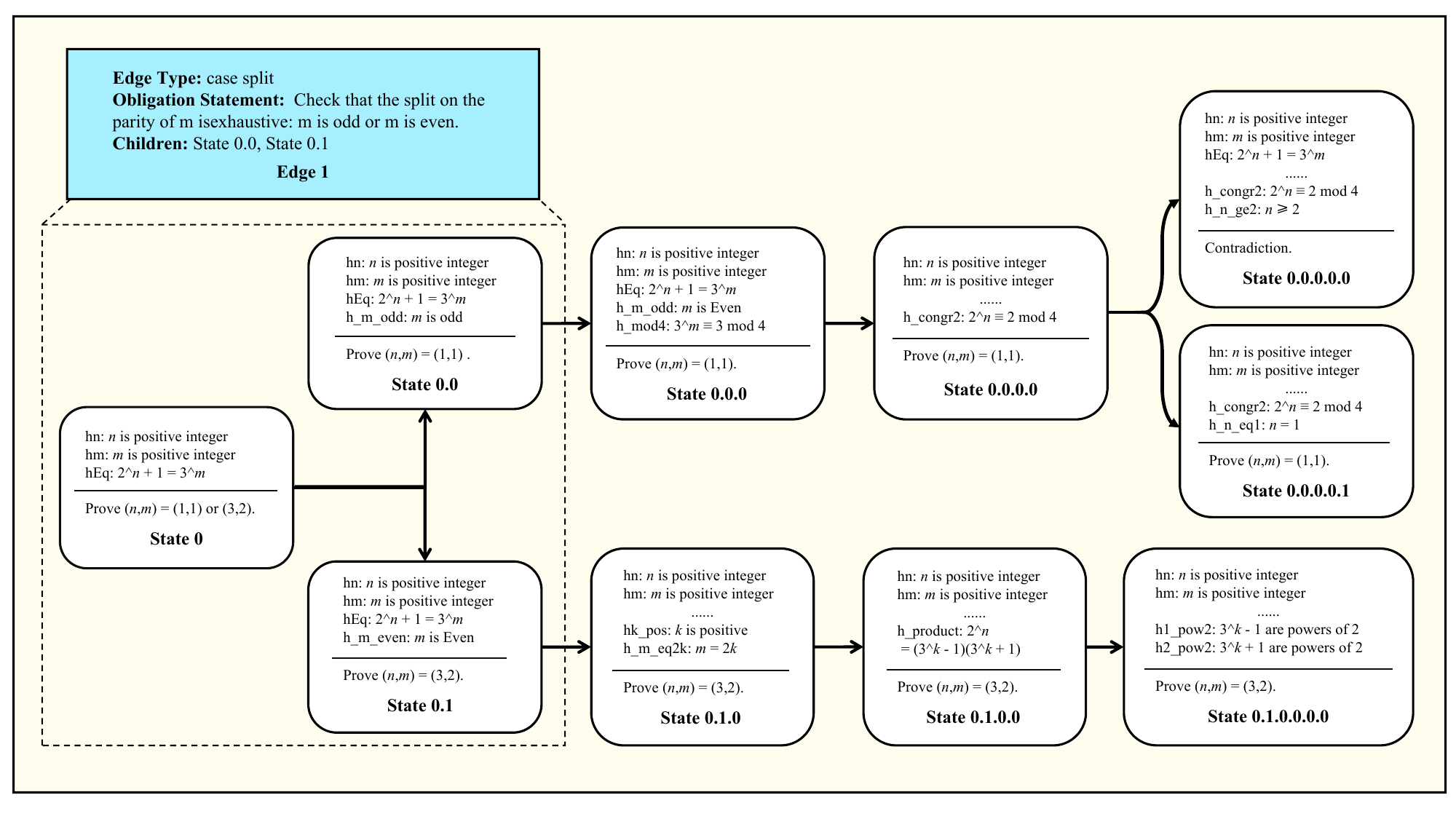}
\caption{From Proof Steps to a Tree of Proof States}
\label{fig:tree-of-states-example}
\end{figure}

\textbf{Definition of a Tree of Proof States.} Each node \(v\in V\) represents a natural-language proof state \(S_v=(\Gamma_v,G_v)\), where \(\Gamma_v\) is the currently available accumulated assumptions and \(G_v\) is the current subgoal to be proved. Taking the first step in Figure~\ref{fig:tree-of-states-example} as an example, after the conditions and conclusion of the original problem in State 0, the natural-language proof wishes to perform a case split on the parity of \(m\). Therefore, State 0.0 adds the branch assumption ``\(m\) is odd'' on top of the parent state and maintains the branch goal \((n,m)=(1,1)\); State 0.1 adds the branch assumption ``\(m\) is even'' and maintains the branch goal \((n,m)=(3,2)\). Together they correspond to the next proof states of State 0 in different proof branches. Each edge \(e=(v\to v')\in E\) represents a local reasoning transition of type \(\tau_e\), such as fact derivation, rewrite, goal reduction, case split, or lemma introduction. Each edge is recorded as an \texttt{EdgeUnit} \(u_e\), serving as the basic object for subsequent review and formal validation. In the example just given, the corresponding edge type is case split.

\textbf{Recursive construction.} In the process of constructing the proof state tree, for each pre-decomposed substep, the model first identifies its transition type \(\tau_e\), then locates the parent state \(v\) where this substep occurs. For different transition types, we have written a detailed state-transition manual to enable the LLM to accurately generate child states \(S_{v'}=(\Gamma_{v'},G_{v'})\). For example, if the substep introduces a new fact \(h\), we update \(\Gamma_{v'} = \Gamma_v \cup \{h\}\); if it changes the current goal, we update \(G_v\); if it produces multiple branches, we generate multiple child nodes with branch-specific conditions. Edge 1 in Figure~\ref{fig:tree-of-states-example} is a case split example: it splits State 0 into State 0.0 and State 0.1, corresponding to \(m\) being odd and \(m\) being even, respectively. We further list state-update templates for different logical patterns in Appendix~\ref{app:tree-of-proof-state}. Thus, recursive construction is not merely text segmentation, but continuously maintains the local proof context carried by each \texttt{EdgeUnit}.

\subsection{Suspicion Tree Search}

After proof-step pre-decomposition and recursive construction of the proof-state tree, we obtain a collection of \texttt{EdgeUnits} that is often much larger than the original set of COT steps. Since Lean validation is expensive, it is impractical to generate formal validation results for every \texttt{EdgeUnit}. We therefore first run Suspicion Search over all edges in the proof-state tree, producing an initial risk tree that estimates which local transitions are more likely to contain the first material error.

Suspicion Search uses the proof-state information attached to each \texttt{EdgeUnit}, including before-context, after-context, goal relation, branch scope, and transition type. The model produces a set of one-shot lightweight judgments: whether the transition is suspicious, how suspicious it is, and an overall suspicion risk score in \([0,1]\). During the search, we provide examples of common failure modes for different edge types and prompt the model to check for the corresponding logical errors. Empirically, more than 80\% of suspicion risk scores fall into the two extreme intervals \([0,0.2]\) and \([0.8,1.0]\). We therefore treat edges with suspicion risk above 0.6 as relatively suspicious candidates.

After scoring all edges, we build an audit window rather than selecting the globally most suspicious \texttt{EdgeUnit}. Among suspicious edges, we choose the earliest one as the focus unit and include a short prefix of preceding edges, especially those with special transition types. This yields a cost-controlled set of local transitions for obligation discovery and validation.

\subsection{Obligation Statement Generation}

\begin{table}[b]
\centering
\caption{Typed obligation kinds for selected \texttt{EdgeUnits}.}
\label{tab:obligation-kinds}
\small
\begin{tabular}{@{}p{0.20\linewidth}p{0.33\linewidth}p{0.36\linewidth}@{}}
\toprule
Type & Local check & Trigger \\
\midrule
Derived fact & New fact follows from context & Implication to new claim \\
Rewrite & Rewrite or calculation is valid & Equivalent rewrite or simplification \\
Goal reduction & After-goal implies before-goal & Sufficient condition for goal \\
Fact plus reduction & New fact and reduction are valid & Claim + goal change \\
Case split & Cases are valid and covering & Case analysis \\
Branch elimination & Discarded branch is impossible & Infeasible branch \\
Witness & Witness satisfies the target & Witness term \\
Contested claim & Flagged claim is true & Flagged claim \\
Entailment & Context entails the claim & Bare assertion \\
Overlap partition & Cases form a proper partition & Partition claim \\
\bottomrule
\end{tabular}
\end{table}

Proof-state tree decomposition gives us local transitions, but a transition itself is not yet a directly checkable mathematical statement. An \texttt{EdgeUnit} records how the proof moves from a before-state to an after-state: available assumptions, newly introduced conditions, goal changes, and branch or source-step scope. To validate such a transition, we summarize its mathematical responsibility as an explicit obligation statement. In most cases, this statement has an implication-like form: under the before-context, a new claim follows; or, under the incoming context, proving the after-goal is sufficient for proving the before-goal.

Concretely, we generate obligations by selecting a conversion template according to the \texttt{EdgeUnit}'s edge type and state change. For common logical actions, we provide examples that map a before-state--after-state transition to an obligation statement, and ask the language model to instantiate the corresponding template for the current edge. The model therefore does not freely summarize the whole proof; it generates a statement matched to the local transition under the given proof states, transition type, new conditions, and goal relation. Table~\ref{tab:obligation-kinds} summarizes the main obligation kinds and their corresponding local checks.

For example, in a \texttt{fact\_plus\_reduction} transition, an edge both introduces a new fact and changes the goal. If we only check whether the after-goal is sufficient for the before-goal, the new claim may be used as a premise before being validated. We therefore split this transition into two obligations: whether the new fact is supported by the prior context, and whether the goal reduction is valid when that fact is available. For a \texttt{case\_split}, the obligation instead checks whether the cases are exhaustive, form a valid partition, or justify the stated branch consequence. Thus, typed conversion turns a local edge into an explicit statement that formal validation can handle.

\section{Faithfulness-Aware Verification and Evidence Fusion}
\label{sec:verification-fusion}

\subsection{Statement Faithfulness Scoring}

Given an obligation \(o\), let \(m\) denote its natural-language statement and \(C\) its local context. Formal validation first generates a Lean statement \(\hat m\). Since Lean validates only \(\hat m\), if \(\hat m\) drops assumptions, changes objects, reverses the implication direction, or weakens the original claim, then a successful Lean proof does not establish the original obligation. Formal validation must therefore be preceded by statement faithfulness checking.

To measure the semantic faithfulness of the formal statement \(\hat m\) to the natural-language obligation \(m\) under the local context \(C\), we define the statement faithfulness score \(S_{\mathrm{faith}}(m,\hat m;C)\in[0,1]\):
\begin{equation}
\label{eq:faithfulness-score}
S_{\mathrm{faith}}(m,\hat m;C)
=
\sqrt{
S_{\mathrm{prem}}(m,\hat m;C)
\cdot
S_{\mathrm{conc}}(m,\hat m;C)
}
\cdot
S_{\mathrm{hol}}(m,\hat m;C).
\end{equation}
Equation~\eqref{eq:faithfulness-score} combines premise fidelity, conclusion fidelity, and a holistic semantic score to measure whether \(\hat m\) still expresses the same local mathematical claim. Here, \(S_{\mathrm{prem}}\) measures whether the hypotheses of \(\hat m\) faithfully retain the local context, branch assumptions, and fixed witnesses needed by the obligation, while \(S_{\mathrm{conc}}\) measures whether the conclusion of \(\hat m\) matches the target claim of \(m\). Let \(T_{\mathrm{prem}}(m,C)\) denote the premise-side semantic slots extracted from the context and the source step, and let \(T_{\mathrm{conc}}(m)\) denote the conclusion-side semantic slots in the obligation. The value \(\mathrm{match}(t,\hat m)\in[0,1]\) is assigned by the semantic checker rather than by string matching: for each semantic slot \(t\), the checker compares the source slot with the hypotheses or conclusion of \(\hat m\), and judges whether it is preserved, omitted, weakened, strengthened, or replaced. Preserved slots receive high scores, while missing critical assumptions, altered objects, or changed claims receive low scores. The resulting slot scores are defined in Equation~\eqref{eq:prem-conc-score}.
{\small
\begin{equation}
\label{eq:prem-conc-score}
S_{\mathrm{prem}}=
\frac{\sum_{t\in T_{\mathrm{prem}}}\mathrm{match}(t,\hat m)}{|T_{\mathrm{prem}}|}
\quad
S_{\mathrm{conc}}=
\frac{\sum_{t\in T_{\mathrm{conc}}}\mathrm{match}(t,\hat m)}{|T_{\mathrm{conc}}|}.
\end{equation}
}
The holistic score \(S_{\mathrm{hol}}\) is determined by five aspects: step-relation fidelity, object and witness fidelity, directionality fidelity, role-alignment fidelity, and syntax fidelity. It captures severe patterns beyond isolated premise or conclusion slots, including wrong direction, role reversal, meta-level wrappers, and vacuous formalizations; details are in Appendix~\ref{app:semantic-checker}.

Together with the faithfulness score, the semantic checker returns a structured semantic report. This report summarizes the diagnostic information produced when the checker evaluates different aspects and semantic elements, including statement-level status, detected drift categories, checker confidence, a natural-language reason, missing assumptions, overgeneralization examples, and a suggested revision. It points out the semantic differences between \(m\) and \(\hat m\): which contextual assumptions are missing, which mathematical objects have been replaced, which implication direction or quantifier structure has changed, and what should be added or revised to repair \(\hat m\).

\subsection{Formal Evaluation Agent}

Using Statement Faithfulness Scoring as a central checking tool, we develop the Formal Evaluation Agent to convert each typed obligation into local checkable evidence. The agent runs on each obligation and returns a local evaluation result containing: (i) the obligation's local checking status in \(\{\texttt{passed},\texttt{refuted},\texttt{inconclusive}\}\), (ii) evidence strength adjusted by faithfulness and checker confidence, and (iii) a checkable artifact or diagnostic message. This result answers only the local obligation; it is not a proof-level verdict.

\textbf{Formal Statement Generation.} First, the statement agent generates a local formal statement \(\hat m\) or a LaTeX mathematical equation from the problem statement, the \texttt{EdgeUnit} summary, and the obligation statement. The latter is generated only when the obligation is a pure mathematical expression. When generating \(\hat m\), the system first asks the LLM for candidate statements and then checks whether the candidate compiles. If it does not compile, the system invokes Lean MCP tools to repair syntax, type, or dependency errors. After compilation succeeds, the semantic checker performs faithfulness scoring between \(\hat m\) and the original obligation \(m\), returning \(S_{\mathrm{faith}}\), component-level semantic judgments, and a structured semantic report. If the language-meaning score or a critical component score falls below the threshold, the agent revises the statement using the drift report and repeats both compilation and semantic checking.

\textbf{Equation SymPy Checker.} We quickly handle pure numeric claims extracted in the previous step, provided they contain neither variables nor natural-language explanation. Concretely, the system normalizes mathematical notation from the LaTeX equation and passes the resulting expression to a SymPy-based numeric checker. This checker handles only pure numeric equality chains and single comparison statements, returning \texttt{passed}, \texttt{refuted}, or \texttt{inconclusive}.

\textbf{Lean Prover Checker.} For other obligations whose statements pass the faithfulness gate, the agent enters the Lean validation branch. According to the suspicion score from the audit stage, the system verifies the original proposition in Lean for obligations that are likely correct, and verifies a negated proposition or the existence of a counterexample for obligations that are likely wrong. A proof-search agent repeatedly interacts with the Lean MCP tools and revises the proof attempt until it succeeds or reaches the time limit. If the original proposition is proved, the result is \texttt{passed}; if the negated proposition or a counterexample is verified, the result is \texttt{refuted}. Timeout or failure to pass the faithfulness gate produces \texttt{inconclusive}, together with a specific inconclusive type.

\textbf{Evidence Fusion and Final Decision.} After obtaining local checking results, the system aggregates evidence back to the original COT steps. Since one \texttt{EdgeUnit} may produce multiple obligations, fusion summarizes local review, typed obligations, formal evaluation results, faithfulness reports, and diagnostics into \(\mathrm{Info}(u_e)\), then maps it to the corresponding step. A step is marked \texttt{incorrect} when it has faithful and reliable negative evidence, such as a refuted obligation, a verified counterexample, or agreement between formal diagnostics and local review; it is marked \texttt{uncertain} when the evidence mainly reflects semantic drift, formalization failure, timeout, or conflicting diagnostics. The system returns the earliest step with reliable negative evidence as the first error; if none is found, the LLM synthesizes all local evidence and outputs \texttt{correct} or a risky step. The complete procedure is shown in Algorithm~\ref{alg:lean-assisted-first-error}.

\begin{algorithm}[!b]
\caption{Lean-assisted first-error localization}
\label{alg:lean-assisted-first-error}
\DontPrintSemicolon
\KwIn{Problem \(p\); natural-language proof \(\pi=(s_1,\ldots,s_n)\)}
\KwOut{\(\hat y\in\{\texttt{correct}\}\cup\{1,\ldots,n\}\)}
\BlankLine
Decompose each coarse step \(s_i\) into faithful fine-grained substeps\;
Build state tree \(\mathcal{T}=(V,E)\) and normalize edges into mapped \texttt{EdgeUnits}\;
Scan \texttt{EdgeUnits} and schedule audit window \(W\) by earliest risk and prefix coverage\;
\ForEach{scheduled \texttt{EdgeUnit} \(u_e\in W\)}{
    Run local natural-language review and compile typed obligations \(O(u_e)\)\;
    \ForEach{obligation \(o\in O(u_e)\)}{
        Generate a Lean statement \(\hat m\) or a LaTeX equation from \(p,u_e,o\)\;
        \uIf{\(o\) is extracted as a pure numeric claim}{
            Normalize the equation and run the SymPy numeric checker\;
        }
        \Else{
            Compile \(\hat m\), repairing syntax, type, or dependency errors if needed\;
            Score faithfulness and return semantic judgments and a structured report\;
            \If{language-meaning or critical component score is low}{
                Revise \(\hat m\) using the drift report and repeat checking\;
            }
            \If{\(\hat m\) still fails the faithfulness gate or checking times out}{
                Return \texttt{inconclusive} with a specific type\;
            }
            \Else{
                Run Lean prover checker on the claim, negation, or counterexample branch\;
            }
        }
        Return local result: \texttt{passed}, \texttt{refuted}, or \texttt{inconclusive}\;
    }
    Aggregate review, obligations, formal results, reports, and diagnostics into \(\mathrm{Info}(u_e)\)\;
}
Map \(\mathrm{Info}(u_e)\) back to original coarse COT steps\;
\uIf{faithful reliable negative evidence is found}{
    Check the preceding uncertain prefix and return the first-error step\;
}
\Else{
    \Return{\texttt{correct} or a risky step after final LLM synthesis}\;
}
\end{algorithm}

\section{Empirical Evaluation}
\label{sec:experiments}

We construct two expert-verified datasets targeting olympiad-style proofs and university-level textbook proofs. The first dataset, \textsc{ProofLoc-Olympiad}, contains 350 algebra and number theory problems. The second dataset, \textsc{ProofLoc-University}, contains 200 problems across topology, linear algebra, abstract algebra, real analysis, convex analysis, and convex optimization. Candidate proofs in both datasets are generated by GPT-4o and then annotated by experts for proof correctness and the first erroneous coarse proof step. Importantly, the ground-truth label is aligned with the benchmark-level proof step, rather than the internal \texttt{EdgeUnit} used by our system.

\subsection{Baselines and System Performance}

We first evaluate the direct-judge baseline. This baseline provides the model with the problem statement and the original natural-language proof, and asks it to output either \texttt{correct} or the first erroneous step in one pass. It does not use the tree of proof states, \texttt{EdgeUnits}, local review, typed obligations, the Formal Evaluation Agent, or Evidence Fusion. Thus, direct judging measures the ability of base models to localize the first error without structured decomposition or formal evidence. Exact and binary metrics are defined in Section~\ref{sec:problem-formulation}.

The base models include GPT-5.4, Gemini-3.1-Pro, Opus-4.6, DeepSeek-v4-Pro, GLM-5.1, and Qwen3.5-9B, covering recent frontier closed models and one 9B-scale open-weight model. Model information follows the corresponding provider model cards, technical reports, or official documentation \citep{openai2026gpt54,google2026gemini31pro,anthropic2026opus46,deepseek2026v4,zhipu2026glm51,qwen2026qwen35}. Table~\ref{tab:direct-base-models} shows strong variation, with GPT-5.4 obtaining the strongest exact accuracy and Qwen3.5-9B serving as the 9B-scale comparison.

\begin{table}[t]
\centering
\begin{minipage}[t]{0.36\linewidth}
\vspace{0pt}
\centering
\caption{\textsc{ProofLoc-University} domain distribution.}
\label{tab:datasets}
\includegraphics[width=\linewidth]{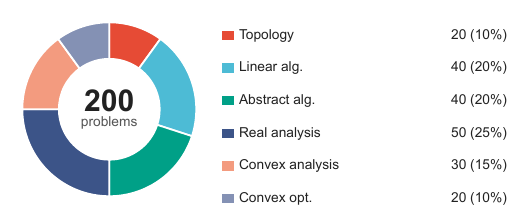}
\end{minipage}
\hfill
\begin{minipage}[t]{0.60\linewidth}
\vspace{0pt}
\centering
\caption{\textsc{ProofLoc-University} domain results.}
\label{tab:university-breakdown}
\footnotesize
\resizebox{\linewidth}{!}{%
\begin{tabular}{@{}l r rr rr@{}}
\toprule
Domain & N & \multicolumn{2}{c}{\textsc{FaithSieve}} & \multicolumn{2}{c}{GPT-5.4 direct} \\
\cmidrule(lr){3-4}\cmidrule(l){5-6}
 & & Exact (\%) & Binary (\%) & Exact (\%) & Binary (\%) \\
\midrule
Overall & 200 & \textbf{84.50} & \textbf{92.50} & 75.00 & 87.00 \\
Topology & 20 & \textbf{80.00} & \textbf{80.00} & 65.00 & 70.00 \\
Linear algebra & 40 & \textbf{100.00} & \textbf{100.00} & 87.50 & 90.00 \\
Abstract algebra & 40 & \textbf{97.50} & \textbf{97.50} & 92.50 & 92.50 \\
Real analysis & 50 & \textbf{80.00} & \textbf{90.00} & 68.00 & 86.00 \\
Convex analysis & 30 & \textbf{60.00} & \textbf{86.67} & 50.00 & 83.33 \\
Convex optimization & 20 & \textbf{80.00} & \textbf{95.00} & \textbf{80.00} & \textbf{95.00} \\
\bottomrule
\end{tabular}%
}
\end{minipage}
\end{table}

\begin{table}[t]
\centering
\caption{Direct-judge base-model results on both datasets.}
\label{tab:direct-base-models}
\footnotesize
\setlength{\tabcolsep}{3.5pt}
\begin{tabular}{@{}llrrrr@{}}
\toprule
\multicolumn{2}{c}{} & \multicolumn{2}{c}{\textsc{ProofLoc-University}} & \multicolumn{2}{c}{\textsc{ProofLoc-Olympiad}} \\
\cmidrule(lr){3-4}\cmidrule(l){5-6}
Base model & Params. & Exact (\%) & Binary (\%) & Exact (\%) & Binary (\%) \\
\midrule
GPT-5.4 & undisclosed & \textbf{75.00} & \textbf{87.00} & \textbf{72.29} & 88.29 \\
Gemini-3.1-Pro & undisclosed & 70.00 & 77.50 & 70.86 & 82.86 \\
Opus-4.6 & undisclosed & 64.50 & 77.00 & 71.71 & \textbf{89.14} \\
DeepSeek-v4-Pro & undisclosed & 72.00 & 82.00 & 48.00 & 71.14 \\
GLM-5.1 & undisclosed & 66.00 & 75.50 & 57.71 & 78.57 \\
Qwen3.5-9B & 9B & 61.00 & 76.50 & 53.71 & 76.00 \\
\bottomrule
\end{tabular}
\end{table}

The complete main-results table is moved to Appendix~\ref{app:full-results}. It reports the performance of \textsc{FaithSieve} under different backbones and compares it with the corresponding direct-judge setting. \textsc{FaithSieve} denotes the full pipeline in Sections~\ref{sec:localizing} and~\ref{sec:verification-fusion}: trees of proof states and \texttt{EdgeUnits}, suspicion-guided audit windows, typed obligation discovery, the Formal Evaluation Agent, statement faithfulness scoring, and Evidence Fusion. With the GPT-5.4 backbone, \textsc{FaithSieve} achieves 81.43\% exact accuracy and 93.71\% binary accuracy on \textsc{ProofLoc-Olympiad}, outperforming GPT-5.4 direct judging at 72.29\% and 88.29\%. On \textsc{ProofLoc-University}, \textsc{FaithSieve} reaches 84.50\% exact / 92.50\% binary accuracy, compared with 75.00\% exact / 87.00\% binary for GPT-5.4 direct judging. With the Qwen3.5-9B backbone, \textsc{FaithSieve} improves over direct judging on both \textsc{ProofLoc-Olympiad} and \textsc{ProofLoc-University}. Table~\ref{tab:university-breakdown} further shows gains in most university domains, while Convex Optimization remains essentially tied.

\subsection{\textsc{FaithSieve} Performance and Ablations}

We evaluate GPT-5.4 and Qwen3.5-9B on both datasets to test the effect of the \textsc{FaithSieve} pipeline. Table~\ref{tab:ablation} reports the \textsc{ProofLoc-Olympiad} results, with experimental settings described in Appendix~\ref{app:pipeline-settings}; the \textsc{ProofLoc-University} results are reported in Appendix~\ref{app:full-results}. Across both datasets and backbones, \textsc{FaithSieve} consistently improves over direct judging, showing that local and faithful formal evidence benefits first-error localization.

We also conduct an ablation study on \textsc{ProofLoc-Olympiad} with GPT-5.4 and Qwen3.5-9B to analyze which components are most important. All ablations use a replay-style setting, where existing intermediate artifacts are fixed and downstream components are replayed; further setup details are provided in Appendix~\ref{app:ablation-definitions}.

\begin{table}[t]
\centering
\caption{Ablation results on \textsc{ProofLoc-Olympiad}.}
\label{tab:ablation}
\footnotesize
\setlength{\tabcolsep}{3.0pt}
\renewcommand{\arraystretch}{0.95}
\begin{tabular}{@{}lrr@{\hspace{1.0em}}lrr@{}}
\toprule
\multicolumn{3}{c}{GPT-5.4} & \multicolumn{3}{c}{Qwen3.5-9B} \\
\cmidrule(lr){1-3}\cmidrule(l){4-6}
Experiment & Exact (\%) & Binary (\%) & Experiment & Exact (\%) & Binary (\%) \\
\midrule
Full \textsc{FaithSieve} & \textbf{81.43} & \textbf{93.71} & Full \textsc{FaithSieve} & \textbf{74.00} & \textbf{89.14} \\
Direct judge & 72.29 & 88.29 & Direct judge & 53.71 & 76.00 \\
Step3 graph judge & 75.71 & 88.57 & Step3 graph judge & 65.71 & 77.71 \\
EdgeUnit + NL only & 73.43 & 88.00 & EdgeUnit + NL only & 71.14 & 86.00 \\
w/o local EdgeUnits & 60.86 & 72.29 & w/o local EdgeUnits & 62.29 & 77.71 \\
w/o semantic gate & 70.86 & 86.29 & w/o semantic gate & 70.86 & 86.00 \\
\bottomrule
\end{tabular}
\end{table}

The Step3 graph judge keeps decomposition and proof-state graph artifacts, but removes local review, Lean validation, and evidence fusion. It reaches 75.71\% exact / 88.57\% binary with GPT-5.4, suggesting that graph organization helps localize erroneous steps even without formal evidence. EdgeUnit + NL only keeps local EdgeUnit review and final synthesis, but removes the Evaluation Agent. It obtains 73.43\% exact / 88.00\% binary with GPT-5.4 and 71.14\% / 86.00\% with Qwen3.5-9B, showing that local EdgeUnit representations are especially helpful for the smaller backbone.

The locality ablation removes local EdgeUnits and falls back to coarse-step-level judging. GPT-5.4 with w/o local EdgeUnits drops to 60.86\% exact / 72.29\% binary, and Qwen3.5-9B with w/o local EdgeUnits obtains 62.29\% exact / 77.71\% binary. This experiment directly supports the locality hypothesis: a benchmark step is an annotation unit, but not necessarily the actual reasoning unit. Without decomposing coarse proof steps into local transitions, the system loses key evidence needed for first-error localization. The semantic-gate ablation relaxes the translation-confidence and statement-confidence thresholds, allowing hybrid-validation results to enter the final evidence set more easily. GPT-5.4 with w/o semantic gate drops to 70.86\% exact / 86.29\% binary; Qwen3.5-9B with w/o semantic gate obtains 70.86\% exact / 86.00\% binary, also below the corresponding Full \textsc{FaithSieve} result. These results show that a Lean or checker success signal is reliable only when the statement remains semantically aligned with the original obligation; without faithfulness filtering, semantic drift contaminates the final judgment.

\FloatBarrier
\section{Conclusion}
\label{sec:conclusion}

This paper studies how formal validation can serve as structured evidence for evaluating natural-language mathematical proofs. Our central point is simple: Lean evidence is most useful when it is local and faithful. \textsc{FaithSieve} operationalizes this view by decomposing coarse steps into \texttt{EdgeUnits}, extracting typed obligations, validating selected obligations with cheap checking and Lean-assisted agents, filtering formal evidence through semantic alignment, and fusing the result into a benchmark-aligned first-error prediction. Across \textsc{ProofLoc-Olympiad} and \textsc{ProofLoc-University}, this local-evidence design improves both correctness judgment and first-error localization over direct judging, and the ablations show that local reasoning units, faithful statement checking, formal or arithmetic evidence, and evidence fusion all contribute. More broadly, \textsc{FaithSieve} provides traceable local evidence for auditing generated proofs, supporting step-level feedback, and exposing where a proof first becomes unreliable; future work should extend this approach to geometry, combinatorics, and definition-heavy university mathematics while improving autoformalization and validation efficiency.

\bibliographystyle{plainnat}
\bibliography{references}

\clearpage
\appendix
\addtocontents{toc}{\protect\setcounter{tocdepth}{2}}
\clearpage
\phantomsection
\renewcommand{\contentsname}{Appendix Contents}
\tableofcontents
\clearpage

\section{Introduction to Lean and Mathlib}
\label{app:lean-mathlib}

Lean is a dependent-type-theoretic interactive theorem prover. A mathematical theorem is first stated as a formal proposition, and a proof is accepted only when Lean's kernel verifies that the submitted proof object has the required type. In practice, proofs are often written in tactic mode: instead of constructing the entire proof term at once, the user applies a sequence of tactics that incrementally transform the current proof situation until no obligation remains.

Mathlib is Lean's community-maintained mathematical library. It provides a large body of definitions, lemmas, theorem statements, notation, tactics, and domain-specific infrastructure. For the purposes of this paper, mathlib largely determines the practical boundary of Lean validation: a natural-language obligation is easier to formalize and check when its objects and background facts can be expressed using existing mathlib definitions and lemmas. Accordingly, Lean evidence in our framework is not merely a consequence of the Lean kernel; it also depends on whether the generated statement faithfully represents the original claim and whether the relevant mathematical vocabulary is available in Lean/mathlib \citep{mathlib2020lean}.

Following the state-and-tactic view developed in prior work on Chain of States for Lean formalization \citep{wang2025chainstates}, a proof state in Lean represents the complete local proof environment at a particular moment. Formally, a state can be viewed as a finite collection of goals,
\[
S=\{G_1,\ldots,G_n\},
\]
where each goal has the form
\[
\{h_{1},\ldots,h_m\}\vdash g.
\]
Here \(\{h_1,\ldots,h_m\}\) is the local hypothesis context of that goal, and \(g\) is the target conclusion that remains to be proved. This is why a proof state captures the full local context of the proof at a given moment: it records not only the theorem being pursued, but also all currently open subgoals, their local assumptions, introduced variables, branch hypotheses, intermediate facts, and remaining targets.

A tactic is an operation that transforms one proof state into another. Applying a tactic may solve a goal, replace it with simpler subgoals, split it into cases, introduce variables or assumptions, rewrite the target, add derived facts to the context, or invoke an existing lemma from mathlib. A tactic proof can therefore be written as a sequence
\[
S_0 \xrightarrow{T_1} S_1 \xrightarrow{T_2} \cdots \xrightarrow{T_k} S_k.
\]
The proof is complete if and only if the final state has no remaining goals. During this process, the number and shape of goals can change substantially: a single goal may split into several branch goals after a case analysis; a difficult target may be reduced to auxiliary lemma subgoals; and solved branches disappear from the state.

This paper's notion of \textbf{Proof State Change} is inspired by this Lean mechanism, but is used at the level of natural-language proof evaluation. We do not assume that every natural-language step has already been formalized as an executable Lean tactic. Instead, we represent a local reasoning transition as a change from a before-state to an after-state. The before-state records the assumptions, facts, branch conditions, and current goal available before the step; the after-state records what the step claims to add, transform, reduce, split, or discharge. A valid local proof step should therefore correspond to a faithful and justified proof-state change: newly introduced facts must follow from the previous context, rewritten expressions must preserve meaning, case splits must cover the original goal, witnesses must satisfy their required properties, and goal reductions must be logically equivalent to, or sufficient for, the original target.

This perspective is central to \textsc{FaithSieve}. Rather than asking Lean or an LLM to judge an entire natural-language proof globally, we decompose the proof into local transitions between proof states. Each transition can then be converted into typed obligations and, when appropriate, checked using Lean/mathlib evidence. Lean proof success is useful only when the generated Lean statement faithfully represents the intended proof-state change; proof failure or timeout is treated as inconclusive rather than as direct evidence that the original proof is wrong.

\section{Local Reasoning Unit Construction}
\label{app:local-reasoning-units}

This section describes the local proof-structure construction used in Section~\ref{sec:localizing}. The pipeline first decomposes a coarse natural-language proof into faithful fine-grained steps, then builds a natural-language proof-state tree, normalizes each tree edge as an \texttt{EdgeUnit}, and finally converts selected units into typed obligations. Decomposition, tree construction, tree suspicion scan, and local review are prompt-driven. Typed obligation discovery is governed by deterministic transition typing and obligation-generation rules; the prompts and algorithmic rules are included below for reproducibility.

\subsection{Faithful Proof Decomposition Prompt}
\label{app:decomposition-prompt}

The decomposition prompt asks the model to split each original proof step into fine-grained substeps while preserving the mathematical content of the original proof. The model is instructed not to repair incorrect equations, missing hypotheses, invalid witnesses, branch assumptions, concrete numerical values, or over-strong conclusions. The stage also keeps pointers from fine-grained steps back to the original coarse step.

\vspace{2mm}
\noindent
\textbf{Prompt for Faithful Proof Decomposition}
\begin{PromptBox}
You are given a mathematical problem and a numbered natural-language proof.
For each original proof step, split it into fine-grained substeps.

Rules:
1. Each substep should contain exactly one logical inference whenever possible.
2. Do not correct mathematical errors.
3. Preserve all equations, inequalities, constants, witnesses, variable scopes,
   and branch assumptions.
4. Preserve decisive words such as only, all, exists, unique, least, greatest,
   must, valid, and sufficient.
5. Remove only motivation-only or restatement-only sentences that introduce no
   new reasoning.
6. If a step contains a substitution, a rewrite, a new fact, and a goal change,
   split them into separate substeps.
7. If a substep is inside a case, contradiction, induction, or witness scope,
   state that scope explicitly.

Output:
For each original step i, return substeps i.1, i.2, ... and keep a pointer to
the original step.
\end{PromptBox}

A repair-prevention check flags strengthening, weakening, dropped decisive words, changed constants, missing branch scope, and action-result splits that make the original inference look more justified than it was.

\subsection{Tree of Proof State}
\label{app:tree-of-proof-state}

The tree of proof states is a natural-language structure inspired by Lean proof states. Each state records accumulated assumptions and a current local goal, while each edge records a reasoning transition. A state describes what is available after a step, not what the step itself says. Goals normally remain the final target except in lemma, witness, induction, or iff subtrees.

\vspace{2mm}
\noindent
\textbf{Prompt for Tree of Informal Proof State}
\begin{PromptBox}
You are a mathematical reasoning expert. Your task is to generate a Tree of
Informal Proof State from a detailed Chain of Thought (CoT).

- Each state contains Conditions and a Goal.
- Conditions are ALL hypotheses/facts available at this point.
- The Goal is the FINAL target to prove, not the intermediate step's conclusion.
- A State represents the accumulated proof context AFTER a reasoning step is
  applied, NOT the step itself.

Condition accumulation:
1. Each child state inherits all conditions from its parent, plus any new
   conditions derived in that step.
2. Give each condition a short name.
3. Some steps transform conditions, such as existential elimination,
   disjunction elimination, and have/lemma introduction.

One-to-one mapping:
EVERY CoT Step MUST correspond to EXACTLY ONE edge in the tree.
Do not skip, merge, or invent CoT-step edges.

For each edge, provide Step Explanation metadata:
Pattern, Goal Relation, Retained Conditions, New Conditions,
Removed/Consumed Conditions, Transformed Conditions, and Explanation.
\end{PromptBox}

\begin{center}
\footnotesize
\begin{tabular}{@{}p{0.24\linewidth}p{0.35\linewidth}p{0.27\linewidth}@{}}
\toprule
Pattern & State update & Goal relation \\
\midrule
Simple implication & Add a newly derived fact while usually keeping the current goal. & \texttt{same\_goal} \\
Case split 2-way & Create two child states with branch-specific assumptions. & \texttt{case\_split} \\
Case split 3+ & Create three or more branch-specific child states. & \texttt{case\_split} \\
Have/lemma & Temporarily create a lemma subgoal, then add the lemma back to the context. & \texttt{lemma\_subgoal} \\
Mathematical induction & Create base and inductive-step states, adding the induction hypothesis in the latter. & \texttt{induction} \\
Existential instantiation & Record a concrete witness and specialize the existential goal to that witness. & \texttt{witness\_instantiation} \\
Existential elimination & Consume an existential condition and add the extracted witness and its property. & \texttt{existential\_elimination} \\
Universal instantiation & Keep the universal statement and add its instance at a specific object. & \texttt{universal\_instantiation} \\
Iff split & Split \(A\leftrightarrow B\) into the two directional subgoals. & \texttt{iff\_split} \\
\bottomrule
\end{tabular}
\end{center}

Every parsed tree edge is normalized as an \texttt{EdgeUnit}. The structure stores identity fields \texttt{unit\_id}, \texttt{edge\_index}, \texttt{parent}, and \texttt{children}; provenance fields \texttt{detailed\_step\_idx}, \texttt{original\_step\_idx}, and \texttt{original\_step\_text}; transition metadata \texttt{pattern}, \texttt{explanation}, and \texttt{goal\_relation}; before/after goals and conditions; delta fields such as \texttt{new\_conditions} and \texttt{new\_condition\_details}; structural cues such as branch scope and consumed or transformed conditions; and scheduling hints such as \texttt{is\_signature}, \texttt{signature\_reasons}, and local risk flags.

In representative Olympiad-style generated examples, a GPT-5.4 \texttt{EdgeUnit} for the step ``By closure under addition, their sum \(u+w\) must lie in \(U\cup W\)'' records the new condition \texttt{h\_sum\_in\_union:} \(u+w\in U\cup W\) while keeping the overall iff goal. A Qwen3.5-9B OlympiadBench example for a complex-number inequality problem produces a noisier same-goal unit whose new condition is a partial answer-path statement about expressing \(|z_1-z_2|\) in terms of \(K\) and solving for the smallest positive integer \(K\). Both examples are informative: clean units expose local mathematical deltas, while malformed units expose structure that downstream review should inspect rather than silently smooth away.

\subsection{Tree-Suspicion Scan and Local Review}
\label{app:tree-suspicion-scan}

The tree scan ranks suspicious units and marks which ones are worth local review or formalization. It is not a final proof verdict. In the experiments, the audit window size is fixed to 6: for a suspicious fine-grained step, the window contains that step together with up to six preceding fine-grained steps. This prefix window is designed for first-error localization, since an error may become salient only after later reasoning has propagated it.

\vspace{2mm}
\noindent
\textbf{System Prompt for Tree-Suspicion Scan}
\begin{PromptBox}
You are a strict mathematical proof auditor.
You are given a proof tree already built from a proof. Your job is NOT to
formalize anything.
Your job is only to rank which edges/segments are most suspicious, with special
emphasis on the earliest plausible error.

Return compact JSON only.
\end{PromptBox}

\vspace{2mm}
\noindent
\textbf{User Prompt Template for Tree-Suspicion Scan}
\begin{PromptBox}
Problem:
{problem_statement}

Original proof:
{original_proof}

Proof-tree units:
{rendered EdgeUnit summaries}

Return JSON with schema:
{
  "overall_verdict": "likely_correct" | "likely_incorrect" | "uncertain",
  "candidates": [
    {
      "unit_id": "edge_0",
      "suspicion": 0.0,
      "likely_wrong": true,
      "should_formalize": true,
      "error_type_prior": ["arithmetic"],
      "reason": "..."
    }
  ]
}

Rules:
- Focus on the earliest plausible error.
- Suspicion must be between 0 and 1.
- Mark should_formalize=true only for units where formal or precise checking
  looks useful.
- If unsure, still provide a candidate list with low-to-medium suspicion values.
\end{PromptBox}

For scheduled units, the local reviewer consumes one unit plus the tree-scan prior and returns a JSON verdict, suspicion score, error type, formalization flag, translation difficulty, and reason.

\subsection{Typed Obligation Discovery}
\label{app:obligation-discovery}

Typed obligation discovery converts a selected local transition into explicit natural-language verification targets. The key rule is that newly introduced facts cannot be used as premises before those facts have themselves been checked.

\vspace{2mm}
\noindent
\textbf{Rule Prompt for Typed Obligation Discovery}
\begin{PromptBox}
Input:
- EdgeUnit u with before-state (Gamma_in |- G_in), after-state (Gamma_out |- G_out),
  transition metadata, new conditions, and source proof text.
- Local review r with verdict, suspicion, error type, should_formalize, and
  translation_difficulty.

Step 1: Infer transition type.
- If text mentions case/split: case_split.
- If text introduces witness/choose/take/let/define: witness.
- If goal changes and new conditions exist: fact_plus_reduction.
- If goal changes without new conditions: goal_reduction.
- If text suggests rewrite/simplify/factor/expand/substitute/evaluate/arithmetic:
  rewrite.
- Otherwise: derived_fact.

Step 2: Generate obligation bundle.
- derived_fact/rewrite: create obligations for the newly introduced claims.
- goal_reduction: check that G_out is sufficient to prove G_in.
- fact_plus_reduction: check the new fact first, then check the reduction.
- witness: check the introduced witness or object.
- case_split: choose case_split, branch_elimination,
  justification_faithfulness, or overlap_partition.

Output fields:
obligation_id, unit_id, kind, statement, reason, use_lean,
translation_confidence, source, lean_priority.
\end{PromptBox}

The obligation kinds include \texttt{derived\_fact}, \texttt{rewrite}, \texttt{goal\_reduction}, \texttt{fact\_plus\_reduction}, \texttt{contested\_claim}, \texttt{entailment}, \texttt{case\_split}, \texttt{branch\_elimination}, \texttt{justification\_faithfulness}, \texttt{overlap\_partition}, and \texttt{witness}. These obligations are still natural-language verification targets; Appendix~\ref{app:evaluation-agent} describes how they are checked by cheap arithmetic routines, Lean validation, semantic alignment, and evidence fusion.

\section{Formal Evaluation Agent}
\label{app:evaluation-agent}

This section expands the Formal Evaluation Agent described in Section~\ref{sec:verification-fusion}. The agent evaluates one typed local obligation at a time and returns local evidence, not a global proof verdict. Its result is one of \texttt{passed}, \texttt{refuted}, or \texttt{inconclusive}, together with a faithfulness score, Lean artifact, diagnostic trace, or semantic-drift report.

\subsection{Cheap Arithmetic Checker}
\label{app:cheap-checker}

The cheap arithmetic checker is a deliberately narrow front-end filter for obligations that are pure numerical claims. It handles numeric equalities, equality chains, and single numeric comparisons. Examples include \(3+4=7\), \(1+2=3=\sqrt{9}\), and \(5<8\). The checker is used only when the extracted claim contains no variables, functions, quantifiers, ellipses, branch conditions, named mathematical objects, or natural-language predicates. This conservative scope is important: the checker is meant to catch obvious arithmetic mistakes at low cost, not to decide symbolic algebraic claims or obligations whose meaning depends on context.

Operationally, the agent normalizes the mathematical surface form into a restricted expression grammar. It accepts numerals, rational constants, decimal constants, parentheses, arithmetic operators, exponentiation with numeric exponent, standard comparisons, and equality-chain separators. LaTeX tokens such as \texttt{\textbackslash frac\{a\}\{b\}}, \texttt{\textbackslash sqrt\{a\}}, \texttt{\textbackslash cdot}, and comparison symbols are normalized before evaluation. The parsed expression is evaluated exactly when possible using rational arithmetic; otherwise the checker falls back to high-precision numeric evaluation with a strict tolerance policy. Equality chains are checked pairwise from left to right. A claim is returned as \texttt{passed} only if all comparisons are verified, \texttt{refuted} if a pure numeric comparison is false, and \texttt{inconclusive} if parsing fails or if the claim falls outside the accepted grammar.

\begin{center}
\begin{tabular}{@{}lll@{}}
\toprule
Claim & Status & Reason \\
\midrule
\(3+4=7\) & \texttt{passed} & Pure numeric equality. \\
\(1^2+2^2=7\) & \texttt{refuted} & False pure numeric equality. \\
\(1+2=3=\sqrt{9}\) & \texttt{passed} & Equality chain. \\
\(5<8\) & \texttt{passed} & Pure numeric comparison. \\
\(\forall x,\ x^2+7\ge 7\) & \texttt{inconclusive} & Variable and quantifier. \\
\(f_1(0)^2+\cdots+f_n(0)^2=7\) & \texttt{inconclusive} & Symbols and ellipsis. \\
\bottomrule
\end{tabular}
\end{center}

\subsection{Statement Semantic Checker}
\label{app:semantic-checker}

The statement semantic checker implements the faithfulness gate in Section~\ref{sec:verification-fusion}. Given a natural-language obligation \(m\), its local context \(C\), and a generated Lean statement \(\hat m\), the checker asks whether \(\hat m\) expresses the same local mathematical claim as \(m\). It does not decide whether the claim is true and does not inspect whether a Lean proof succeeds. Its role is to prevent a provable but semantically shifted Lean theorem from being treated as evidence for the original obligation.

\paragraph{Semantic slots and \(\mathrm{match}(t,t')\).}
The checker first extracts semantic slots from the source obligation and context. Premise-side slots \(T_{\mathrm{prem}}(m,C)\) include local assumptions, branch hypotheses, declared domains, fixed variables, nonzero or positivity conditions, and witnesses that the obligation is allowed to use. Conclusion-side slots \(T_{\mathrm{conc}}(m)\) include the claimed equality, inequality, implication, set relation, existence claim, uniqueness claim, goal reduction, or branch-coverage target.

The generated Lean statement \(\hat m\) is also rendered into a structured semantic view. Let \(T(\hat m)\) be the slots extracted from \(\hat m\), partitioned into Lean premise slots and Lean conclusion slots. For each source slot \(t\), the checker selects the best corresponding Lean slot
\[
t^\star(t,\hat m)=\arg\max_{t'\in T(\hat m)} \mathrm{match}(t,t'),
\]
where the candidate set is restricted to premise slots when \(t\in T_{\mathrm{prem}}\) and to conclusion slots when \(t\in T_{\mathrm{conc}}\), except that the checker may explicitly flag a role swap if a source premise appears only as a Lean conclusion or vice versa.

The slot-level score \(\mathrm{match}(t,t')\in[0,1]\) is assigned by the following rubric:
\begin{center}
\begin{tabular}{@{}p{0.12\linewidth}p{0.78\linewidth}@{}}
\toprule
Score & Meaning \\
\midrule
1.00 & Exact semantic preservation: same objects, domains, relations, quantifiers, and role. \\
0.75 & Minor notation or definitional variation, with no mathematical weakening or strengthening. \\
0.50 & Partial preservation: main object present, but a modifier, side condition, or scope is missing. \\
0.25 & Weak semantic overlap: related symbols appear, but the claim, role, or relation is changed. \\
0.00 & Missing, contradicted, role-swapped, wrong object, wrong direction, or unrelated formalization. \\
\bottomrule
\end{tabular}
\end{center}
When no suitable Lean slot exists, \(\mathrm{match}(t,\bot)=0\). If a slot is strengthened or weakened, the score depends on whether the change affects the evidential use of the Lean result. For example, adding an extra assumption to a Lean theorem can make the statement easier and therefore less faithful, while dropping a positivity condition can make an originally valid local inference invalid.

The premise and conclusion scores are
\[
S_{\mathrm{prem}}(m,\hat m;C)
=
\frac{1}{|T_{\mathrm{prem}}(m,C)|}
\sum_{t\in T_{\mathrm{prem}}(m,C)}
\max_{t'\in T_{\mathrm{prem}}(\hat m)} \mathrm{match}(t,t'),
\]
\[
S_{\mathrm{conc}}(m,\hat m;C)
=
\frac{1}{|T_{\mathrm{conc}}(m)|}
\sum_{t\in T_{\mathrm{conc}}(m)}
\max_{t'\in T_{\mathrm{conc}}(\hat m)} \mathrm{match}(t,t').
\]
If a slot set is empty because the obligation has no explicit premises or no explicit conclusion-side modifiers, the corresponding average is defined over a singleton vacuous slot with score 1, unless the checker detects that the formal statement introduced unsupported structure. The final faithfulness score is the one used in Section~\ref{sec:verification-fusion}:
\[
S_{\mathrm{faith}}(m,\hat m;C)
=
\sqrt{
S_{\mathrm{prem}}(m,\hat m;C)
\cdot
S_{\mathrm{conc}}(m,\hat m;C)}
\cdot
S_{\mathrm{hol}}(m,\hat m;C).
\]

\paragraph{Five-aspect holistic scoring prompt.}
The holistic score \(S_{\mathrm{hol}}\) is produced by a five-aspect semantic scoring prompt. Each aspect is scored in \(\{0,0.25,0.5,0.75,1\}\), and the final holistic value is
\[
S_{\mathrm{hol}}
=
0.20S_{\mathrm{step}}
+0.20S_{\mathrm{obj}}
+0.20S_{\mathrm{dir}}
+0.20S_{\mathrm{role}}
+0.20S_{\mathrm{syn}}.
\]
The uniform weighting treats the five holistic aspects as equally necessary safeguards: even if the same mathematical expression appears, the statement is not faithful if it certifies the wrong transition, changes objects, reverses direction, swaps logical roles, or collapses into a vacuous surface form.

\vspace{2mm}
\noindent
\textbf{Prompt for Statement Semantic Checking}
\begin{PromptBox}
System:
You are a strict semantic checker for Lean autoformalization.
Your task is not to prove the statement. Your task is to judge whether the
Lean statement faithfully represents the natural-language local obligation
under the given context.

Input:
1. Problem statement.
2. Source proof step and local EdgeUnit context.
3. Natural-language obligation m.
4. Generated Lean statement m_hat.
5. Optional Lean diagnostics from compilation.

Score each aspect using only {0, 0.25, 0.5, 0.75, 1}.

Aspect A: step_relation_fidelity.
Does m_hat formalize the same local proof-state transition as m?
Check whether it validates the same newly introduced fact, rewrite,
goal reduction, witness check, case split, or branch elimination.

Aspect B: object_witness_fidelity.
Are the mathematical objects, domains, constants, functions, sets,
indices, witnesses, and branch-specific variables the same?

Aspect C: directionality_fidelity.
Are implication direction, iff direction, equality orientation,
goal-reduction direction, quantifier order, and negation polarity preserved?

Aspect D: role_alignment_fidelity.
Are assumptions, conclusions, definitions, intermediate claims, and goals
kept in the same logical roles?

Aspect E: syntax_surface_fidelity.
Is the formal statement a direct mathematical statement rather than a
vacuous wrapper, placeholder, trivial theorem, malformed parse, or
meta-level assertion?

Return compact JSON:
{
  "scores": {
    "step_relation_fidelity": 0.0,
    "object_witness_fidelity": 0.0,
    "directionality_fidelity": 0.0,
    "role_alignment_fidelity": 0.0,
    "syntax_surface_fidelity": 0.0
  },
  "holistic_score": 0.0,
  "status": "faithful" | "repairable_drift" | "unfaithful",
  "drift_categories": [
    "missing_assumptions",
    "overgeneralized",
    "undergeneralized",
    "wrong_objects",
    "wrong_quantifiers",
    "wrong_direction",
    "role_swap",
    "vacuous_or_trivialized"
  ],
  "reason": "...",
  "missing_or_changed_slots": ["..."],
  "suggested_revision": "..."
}
\end{PromptBox}

The checker returns \texttt{faithful} only when the total faithfulness score and all critical component scores exceed the threshold. Directionality, role alignment, and conclusion fidelity are treated as critical: a high average score cannot compensate for proving the converse, assuming the target, or changing the contested conclusion.

\subsection{Statement / Proof Generation and Compile Repair}
\label{app:statement-proof-generation-repair}

For obligations that are not discharged by the cheap arithmetic checker, the Evaluation Agent first generates Lean statements and, when appropriate, then generates Lean proofs. The agent interacts with Lean through Lean Tools MCP \citep{leanToolsMCP2026}, using the returned compilation diagnostics, goal states, and proof-attempt results as feedback for repair. The key point here is not the MCP interface itself, but the agentic loop that separates statement generation, semantic checking, and proof generation.

The statement and proof stages are separated. A compiled Lean statement is not enough, because it may be semantically unfaithful. Conversely, an unproved Lean statement is not evidence that the natural-language obligation is false, since proof search may fail for engineering reasons. The agent first searches for a compilable and faithful statement; only after the semantic gate is passed does it enter proof generation.

The statement-generation loop alternates between Lean compilation and semantic checking. If Lean reports a syntax, type, import, dependency, namespace, or elaboration error, the next prompt contains the current candidate and MCP diagnostics, and the model is asked to repair only the Lean artifact. If the candidate compiles, the semantic checker compares it against \(m\) under \(C\). When semantic drift is detected, the structured semantic report is appended to the next statement-generation prompt, and the agent revises the formal statement while preserving the original obligation. The loop terminates when a candidate both compiles and passes the faithfulness gate, or when the budget is exhausted.

The proof-generation loop starts only from a compiled and faithful statement. The proof agent proposes a Lean proof script, checks it through MCP, and repairs only the proof body according to Lean diagnostics and current goal states. In this stage the statement is held fixed unless the system explicitly returns to statement generation after a confirmed statement-level issue. If proof succeeds, the local result is \texttt{passed} for the original-claim branch. If the system is running a negation or counterexample branch and that branch is verified, the local result is \texttt{refuted}. Otherwise, timeout, search exhaustion, or unresolved compiler errors produce \texttt{inconclusive}.

\begin{algorithm}[!t]
\caption{Statement and proof generation with MCP compile repair}
\label{alg:statement-proof-repair}
\DontPrintSemicolon
\KwIn{Obligation \(o=(\kappa,m,C,u_e,\rho)\); problem \(p\); statement budget \(B_s\); proof budget \(B_p\); faithfulness threshold \(\tau\)}
\KwOut{Local result in \(\{\texttt{passed},\texttt{refuted},\texttt{inconclusive}\}\)}
\If{\(m\) is eligible for cheap arithmetic checking}{
    \Return{\textsc{CheapArithmeticChecker}(\(m\))}\;
}
\(\textit{prompt}\leftarrow\textsc{BuildStatementPrompt}(p,C,o)\)\;
\(\textit{report}\leftarrow\emptyset\)\;
\For{\(i=1,\ldots,B_s\)}{
    \(\hat m\leftarrow\textsc{GenerateLeanStatement}(\textit{prompt},\textit{report})\)\;
    \(\textit{cr}\leftarrow\textsc{MCP.lean\_run\_code}(\hat m)\)\;
    \uIf{\(\textit{cr}\) has Lean errors}{
        \(\textit{diag}\leftarrow\textsc{MCP.lean\_diagnostic\_messages}(\hat m)\)\;
        \(\textit{prompt}\leftarrow\textsc{AddCompileDiagnostics}(\textit{prompt},\textit{diag})\)\;
    }
    \Else{
        \(\textit{report}\leftarrow\textsc{SemanticChecker}(m,\hat m,C)\)\;
        \uIf{\(\textit{report}.S_{\mathrm{faith}}\ge\tau\) and no critical drift}{
            \textbf{goto} proof stage\;
        }
        \Else{
            \(\textit{prompt}\leftarrow\textsc{AddSemanticReport}(\textit{prompt},\textit{report})\)\;
        }
    }
}
\Return{\texttt{inconclusive} with statement-generation diagnostics}\;
\BlankLine
\textbf{proof stage:}\;
\(\textit{branch}\leftarrow\textsc{SelectBranch}(o,\textit{suspicion prior})\)\;
\(\textit{pprompt}\leftarrow\textsc{BuildProofPrompt}(p,C,o,\hat m,\textit{branch})\)\;
\For{\(j=1,\ldots,B_p\)}{
    \(\pi\leftarrow\textsc{GenerateLeanProof}(\textit{pprompt})\)\;
    \(\textit{pr}\leftarrow\textsc{MCP.lean\_run\_code}(\hat m,\pi)\)\;
    \uIf{\(\textit{pr}\) succeeds}{
        \uIf{\(\textit{branch}=\texttt{original}\)}{
            \Return{\texttt{passed} with Lean artifact and faithfulness score}\;
        }
        \Else{
            \Return{\texttt{refuted} with Lean artifact and faithfulness score}\;
        }
    }
    \(\textit{diag}\leftarrow\textsc{MCP.lean\_diagnostic\_messages}(\pi)\)\;
    \(\textit{goals}\leftarrow\textsc{MCP.lean\_goal}\) when available\;
    \(\textit{pprompt}\leftarrow\textsc{AddProofDiagnostics}(\textit{pprompt},\textit{diag},\textit{goals})\)\;
}
\Return{\texttt{inconclusive} with proof-search diagnostics}\;
\end{algorithm}

This two-loop design keeps failed Lean interactions conservative. Statement compilation errors are treated as formalization failures; semantic drift is treated as an unfaithful artifact; proof failure is treated as insufficient formal evidence. Only a proof or refutation of a statement that has passed the faithfulness gate becomes strong local evidence for first-error localization.

\section{ProofLoc Dataset}
\label{app:proofloc-dataset}

\subsection{Dataset Summary and Source Provenance}
\label{app:proofloc-summary}

This section supplements the dataset discussion in Section~\ref{sec:experiments}. We evaluate on two expert-verified benchmarks. \textsc{ProofLoc-Olympiad} contains 350 olympiad-style algebra and number-theory problems. \textsc{ProofLoc-University} contains 200 university-level problems across topology and metric spaces, linear algebra, abstract algebra, real analysis, convex analysis, and convex optimization.

In both datasets, candidate proofs are generated as numbered step-by-step natural-language proofs. Gold labels are expert judgments aligned to benchmark-level coarse proof steps rather than to the internal \texttt{EdgeUnits} used by \textsc{FaithSieve}. Table~\ref{tab:proofloc-university-provenance} reports the textbook provenance for \textsc{ProofLoc-University}, making the six-domain coverage auditable.

\begin{table}[h]
\centering
\caption{Textbook provenance of \textsc{ProofLoc-University}.}
\label{tab:proofloc-university-provenance}
\footnotesize
\resizebox{\linewidth}{!}{%
\begin{tabular}{@{}lrl l@{}}
\toprule
Domain & \(N\) & Source book & Author(s) \\
\midrule
Topology / metric spaces & 20 & \emph{Topology}~\citep{munkres2000topology} & James R. Munkres \\
Linear algebra & 40 & \emph{Linear Algebra Done Right}~\citep{axler2015linear} & Sheldon Axler \\
Abstract algebra & 40 & \emph{Abstract Algebra}~\citep{dummit2004abstract} & David S. Dummit; Richard M. Foote \\
Real analysis & 50 & \emph{Analysis II}~\citep{tao2016analysisii} & Terence Tao \\
Convex analysis & 30 & \emph{Convex Analysis}~\citep{rockafellar1970convex} & R. Tyrrell Rockafellar \\
Convex optimization & 20 & \emph{Convex Optimization}~\citep{boyd2004convex} & Stephen Boyd; Lieven Vandenberghe \\
\bottomrule
\end{tabular}%
}
\end{table}

\subsection{Data Generation}
\label{app:proofloc-generation}

Candidate proofs are generated by GPT-4o as numbered step-by-step proofs. The generation prompt asks the model to produce a complete proof, not to intentionally produce an incorrect solution. Thus, the errors in the benchmark arise from naturally occurring model reasoning failures and are then identified by expert annotation. This setting is intended to capture proof-step errors that occur during ordinary proof generation, rather than errors produced by adversarial perturbation.

Each generated proof is annotated by mathematics Ph.D. students from well-known universities. Annotators are instructed to judge proof correctness at the benchmark step level and to mark the earliest step that introduces a material proof break; correct proofs are labeled as \texttt{correct}. When annotators encounter disagreement or ambiguity, the example is not resolved by a single unilateral label. Instead, the annotators jointly review the problem statement, the generated proof, the candidate first-error positions, and the downstream dependence of later steps on the disputed claim. The final label is assigned only after this adjudication process reaches agreement on whether the disputed step is mathematically recoverable or constitutes the first unrepaired error in the proof.

\vspace{2mm}
\noindent
\textbf{Proof Generation Prompt}
\begin{PromptBox}
Solve the following mathematical problem.
Write a numbered, step-by-step proof.
Each step should contain a clear mathematical inference.
Do not skip important reasoning steps.

Problem:
{problem}
\end{PromptBox}

\section{Human Audits of Pipeline Components}
\label{app:human-audits-section}

\ifdefined\arxivpublic
  \subsection{Human Audits of Local Reasoning Construction}
\label{app:human-audits}

\paragraph{Decomposition preservation.}
We audited 50 original proof-decomposition logs to test whether the
fine-grained decomposition preserved the first error in the source proof.  Two
independent human experts found preservation in 42/50 (84\%) and 43/50 (86\%)
proofs, respectively.  In their initial judgments, they agreed that 41
decompositions were fully reliable and that 6 contained partial decomposition
issues.  After jointly reviewing the remaining disagreements and the location
of each issue, the consolidated result was 43/50 first-error preservation.
Some decomposition defects occurred only after the source first error and
therefore did not alter the final first-error location.

\paragraph{Suspicion Search calibration.}
We used a pre-frozen sample of 50 Olympiad proofs, containing 25 correct and 25
known-wrong proofs and 752 \texttt{EdgeUnits}.  One independent human reviewer
blindly labeled each proof and unit without access to Search selections,
benchmark labels, Lean/gate results, or final predictions.  After excluding
three uncertain unit labels, Search obtained the results in
Table~\ref{tab:search-human-audit}.

\begin{table}[h]
\centering
\caption{Human calibration of Suspicion Search.}
\label{tab:search-human-audit}
\footnotesize
\begin{tabular}{@{}lrr@{}}
\toprule
Audit measure & Count & Result \\
\midrule
EdgeUnit precision & $174/(174+57)$ & 75.32\% \\
EdgeUnit recall & $174/(174+93)$ & 65.17\% \\
Known-wrong first-error coverage & $23/25$ & 92.00\% \\
Conditional first-error recall & $23/24$ & 95.83\% \\
\bottomrule
\end{tabular}
\end{table}

The unit-level TP/TN/FP/FN counts were 174/425/57/93.  Although the unit-level
recall is moderate, the scheduling objective is proof-level coverage of the
earliest error: Search covered that error in 23 of 25 known-wrong proofs.

  \subsection{Human Calibration of the Semantic Gate}
\label{app:semantic-gate-audit}

Two independent human experts blindly reviewed 135 evaluable theorem/lemma
statements from historical execution records.  Each generated statement was
compared with its natural-language obligation and local context, and
disagreements were resolved before computing the confusion matrix in
Table~\ref{tab:semantic-gate-human-audit}.

\begin{table}[h]
\centering
\caption{Human calibration of the statement semantic gate on 135 statements.}
\label{tab:semantic-gate-human-audit}
\footnotesize
\begin{tabular}{@{}lrr@{}}
\toprule
Human judgment & Gate accept & Gate reject \\
\midrule
Faithful & 105 & 10 \\
Unfaithful & 13 & 7 \\
\bottomrule
\end{tabular}
\end{table}

Treating a faithful statement as positive and gate acceptance as the predicted
positive, the gate achieved 88.98\% precision and 91.30\% recall.  Its FPR was
65.00\% and its FNR was 8.70\%.  The 13 accepted-drift cases had overlapping
failure modes: added or strengthened assumptions (5), vacuous or trivial
propositions (6), and weakened targets (5).  These results show that the gate
substantially agrees with human judgments but is not itself a fidelity
guarantee; accepted formal evidence must remain traceable to its source
obligation.

\fi

\section{Additional Experimental Details and Ablations}
\label{app:experimental-details}

\ifdefined\arxivpublic
  \subsection{Tree Scale and Scheduling Sparsity}
\label{app:tree-scale-scheduling}

On the complete Olympiad-350 artifacts, Suspicion Search reduces the set sent
to expensive downstream checks as shown in Table~\ref{tab:tree-scale}.

\begin{table}[h]
\centering
\caption{Tree and candidate sizes on the complete Olympiad-350 artifacts.}
\label{tab:tree-scale}
\footnotesize
\begin{tabular}{@{}lrrr@{}}
\toprule
Structure & Mean & Median & 90th percentile \\
\midrule
Raw proof-state nodes & 16.81 & 15 & 28 \\
Normalized non-placeholder EdgeUnits & 15.59 & 14 & 27 \\
Search-selected candidates & 4.91 & 5 & 7 \\
\bottomrule
\end{tabular}
\end{table}

Overall, 1,720 of 5,458 EdgeUnits (31.51\%) were selected as candidates, and
formalization was restricted to this candidate set.  The prefix-lookback
parameter was fixed to 6: each local audit receives the focus unit and up to
six preceding fine-grained units, for a maximum contiguous context of seven
units.  This parameter is configurable and is not selected by the LLM.

\fi

\subsection{Direct Judge Baseline}
\label{app:direct-prompt}

The direct judge baseline is the most direct natural-language judging baseline in our experiments. It receives only the problem statement and the original numbered proof steps, and is asked to output \texttt{correct} or the earliest erroneous step number. It does not use proof decomposition, proof-state graphs, \texttt{EdgeUnits}, local review, cheap arithmetic checking, Lean validation, semantic gates, or evidence fusion. This baseline measures the ability of a base model to perform proof correctness judgment and first-error localization without structured local evidence.

Each solution is formatted as numbered steps before being inserted into the user message, using the form \texttt{Step 1: ...}, \texttt{Step 2: ...}. The model output is restricted to \texttt{correct} or \texttt{step N}. During evaluation, \texttt{correct} is parsed as label 0 and \texttt{step N} is parsed as the first erroneous step \(N\); unparsable outputs are counted as parse failures.

\vspace{2mm}
\noindent
\textbf{Direct Judge Baseline Prompt}
\begin{PromptBox}
System prompt:
You are a mathematical reasoning evaluator. Given a math problem and a
step-by-step solution, your task is to identify whether the solution contains
an error and, if so, which step first introduces the error.

Rules:
- Carefully check each step for mathematical correctness.
- A step is incorrect if it contains a wrong equation, invalid logical
  deduction, unjustified claim, or a conclusion that does not follow from
  previous steps.
- Report the FIRST step where an error appears.
- If all steps are correct, report "correct".

Output format (respond with ONLY one of these, nothing else):
- If all steps are correct: correct
- If step N is the first error: step N

User prompt template:
Problem:
{problem}

Solution steps:
{steps}

Identify the first error. Respond with ONLY "correct" or "step N"
(where N is the step number).
\end{PromptBox}

\subsection{FaithSieve Pipeline Settings}
\label{app:pipeline-settings}

The full \textsc{FaithSieve} pipeline includes proof decomposition, proof-state graph construction, \texttt{EdgeUnit} construction, suspicion-guided scheduling, typed obligation discovery, EvaluationAgent validation, and evidence fusion. Different backbones use the same task definition, the same benchmark labels, and the same final output format: \texttt{correct} or the earliest erroneous benchmark-level proof step. Since benchmark labels are attached to the original proof steps rather than to internal \texttt{EdgeUnits} or Lean obligations, the final stage maps obligation-level evidence back to the original step index.

All LLM calls are issued through OpenAI-compatible APIs or the corresponding provider APIs. Closed or hosted frontier models are queried through their provider endpoints. The experimental pipeline does not use an external coding-agent framework; models receive prompts through APIs and return text responses. Lean file creation, file editing, snippet checking, diagnostics, proof-state inspection, and timeout-controlled validation are performed through \texttt{lean-tools-MCP}.

Lean validation is performed on CPU. For each local obligation sent to the Formal Evaluation Agent, the system first generates a Lean statement and checks whether it compiles. Only after the statement passes the semantic gate does the system attempt proof, refutation, or counterexample search. The Lean proof loop uses a 10-minute timeout. This threshold is intentionally permissive because the checked targets are local obligations produced after decomposition, such as algebraic rewrites, derived facts, goal reductions, witness checks, case-split subgoals, or simple entailments. A timeout is treated as inconclusive formal evidence rather than as direct evidence that the original proof step is wrong.

\ifdefined\arxivpublic
\else
\textbf{An anonymized repository is available at} \url{https://anonymous.4open.science/r/anonymous-faithsieve-8662/}, where code and datasets will be uploaded during the anonymous review period.
\fi

\subsection{Full Main Results}
\label{app:full-results}

Table~\ref{tab:faithsieve-main-app} reports the full main results corresponding to Section~\ref{sec:experiments}. Exact accuracy requires the predicted first erroneous step to match the benchmark label exactly, while binary accuracy evaluates only the correct-versus-incorrect decision. The \textsc{FaithSieve} rows in this table denote complete end-to-end pipeline runs.

\begin{table}[h]
\centering
\caption{Main \textsc{FaithSieve} results.}
\label{tab:faithsieve-main-app}
\footnotesize
\resizebox{\linewidth}{!}{%
\begin{tabular}{@{}l rr@{\hspace{1.0em}}l rr@{}}
\toprule
\multicolumn{3}{c}{\textsc{ProofLoc-Olympiad}} & \multicolumn{3}{c}{\textsc{ProofLoc-University}} \\
\cmidrule(lr){1-3}\cmidrule(l){4-6}
Setting & Exact (\%) & Binary (\%) & Setting & Exact (\%) & Binary (\%) \\
\midrule
Direct (GPT-5.4) & 72.29 & 88.29 & Direct (GPT-5.4) & 75.00 & 87.00 \\
\textsc{FaithSieve} (GPT-5.4) & \textbf{81.43} & \textbf{93.71} & \textsc{FaithSieve} (GPT-5.4) & \textbf{84.50} & \textbf{92.50} \\
Direct (Qwen3.5-9B) & 53.71 & 76.00 & Direct (Qwen3.5-9B) & 61.00 & 76.50 \\
\textsc{FaithSieve} (Qwen3.5-9B) & \textbf{74.00} & \textbf{89.14} & \textsc{FaithSieve} (Qwen3.5-9B) & \textbf{69.50} & \textbf{78.50} \\
\bottomrule
\end{tabular}%
}
\end{table}

\subsection{Ablation Study and Analysis}
\label{app:ablation-definitions}

The ablation experiments evaluate the contribution of individual components by preserving or removing specific parts of the \textsc{FaithSieve} pipeline. Whenever possible, the experiments reuse intermediate artifacts produced by the full pipeline and replay the downstream synthesis or validation stages affected by the ablated component. This design makes the ablations closer to component diagnosis: the comparison asks what changes when a specific source of evidence is removed or weakened, while reducing variation from rerunning earlier LLM-based stages.

Table~\ref{tab:ablation-components-app} summarizes the component configuration of each ablation variant, and Table~\ref{tab:ablation-app} reports the corresponding results on \textsc{ProofLoc-Olympiad}. The table omits an interpretation column so that the numeric comparison remains explicit; the following paragraphs discuss the mechanism tested by each variant.

\begin{table}[h]
\centering
\caption{Component configuration of ablation variants.}
\label{tab:ablation-components-app}
\footnotesize
\resizebox{\linewidth}{!}{%
\begin{tabular}{@{}lcccccc@{}}
\toprule
Variant & Decomp./graph & \texttt{EdgeUnits} & Local NL review & Cheap/Lean evidence & Semantic gate & Evidence fusion \\
\midrule
Direct judge & No & No & No & No & No & No \\
Step3 graph judge & Yes & Graph-level only & No & No & No & No \\
EdgeUnit + NL only & Yes & Yes & Yes & No & No formal gate & Yes \\
w/o local EdgeUnits & Yes & No / coarse-step & Partial & Yes & Yes & Yes \\
w/o semantic gate & Yes & Yes & Yes & Yes & Disabled or loosened & Yes \\
Full \textsc{FaithSieve} & Yes & Yes & Yes & Yes & Yes & Yes \\
\bottomrule
\end{tabular}%
}
\end{table}

\begin{table}[h]
\centering
\caption{Ablation results on \textsc{ProofLoc-Olympiad}.}
\label{tab:ablation-app}
\footnotesize
\setlength{\tabcolsep}{3.5pt}
\begin{tabular}{@{}lrr@{\hspace{1.0em}}lrr@{}}
\toprule
\multicolumn{3}{c}{GPT-5.4} & \multicolumn{3}{c}{Qwen3.5-9B} \\
\cmidrule(lr){1-3}\cmidrule(l){4-6}
Variant & Exact (\%) & Binary (\%) & Variant & Exact (\%) & Binary (\%) \\
\midrule
Full \textsc{FaithSieve} & \textbf{81.43} & \textbf{93.71} & Full \textsc{FaithSieve} & \textbf{74.00} & \textbf{89.14} \\
Direct judge & 72.29 & 88.29 & Direct judge & 53.71 & 76.00 \\
Step3 graph judge & 75.71 & 88.57 & Step3 graph judge & 65.71 & 77.71 \\
EdgeUnit + NL only & 73.43 & 88.00 & EdgeUnit + NL only & 71.14 & 86.00 \\
w/o local EdgeUnits & 60.86 & 72.29 & w/o local EdgeUnits & 62.29 & 77.71 \\
w/o semantic gate & 70.86 & 86.29 & w/o semantic gate & 70.86 & 86.00 \\
\bottomrule
\end{tabular}
\end{table}

The direct-judge baseline removes the entire structured-evidence pipeline. It reads the problem statement and the original numbered proof steps, then predicts \texttt{correct} or the first erroneous step in a single pass. GPT-5.4 direct judging obtains 72.29\% exact / 88.29\% binary accuracy, while Qwen3.5-9B obtains 53.71\% exact / 76.00\% binary accuracy. The gap to the full setting indicates that first-error localization benefits from organizing the proof into local evidence-bearing units rather than relying only on whole-proof natural-language judgment.

The Step3 graph judge keeps the decomposition and proof-state graph artifacts, but removes local EdgeUnit review, typed obligation discovery, Lean validation, and final evidence fusion. This variant tests whether tree-level organization alone can improve localization. It improves over direct judging for both backbones, reaching 75.71\% exact / 88.57\% binary with GPT-5.4 and 65.71\% exact / 77.71\% binary with Qwen3.5-9B. The improvement is larger for exact accuracy than for binary accuracy, suggesting that the graph mainly helps identify where the reasoning first goes wrong, while still lacking the local verification evidence needed by the full system.

The EdgeUnit + NL only variant keeps local EdgeUnit-level review and final synthesis, but disables the Formal Evaluation Agent, including cheap arithmetic checking, Lean statement generation, Lean validation, and the semantic gate for formal evidence. With GPT-5.4 it obtains 73.43\% exact / 88.00\% binary accuracy; with Qwen3.5-9B it obtains 71.14\% exact / 86.00\% binary accuracy. The strong improvement for Qwen3.5-9B shows that local EdgeUnit representations help a smaller backbone focus on checkable transitions even without formal validation, although this setting remains below the full replay result.

The locality ablation removes local EdgeUnits and falls back to coarse-step-level evidence units. GPT-5.4 drops to 60.86\% exact / 72.29\% binary accuracy, and Qwen3.5-9B obtains 62.29\% exact / 77.71\% binary accuracy. This is the largest degradation among the reported variants. The result supports the locality hypothesis: benchmark steps are annotation units, but they are often too coarse to serve as reliable reasoning units. Decomposing them into local transitions is therefore essential for detecting the first material proof break.

The semantic-gate ablation keeps hybrid validation but disables or loosens the statement-faithfulness filter before formal evidence is admitted into the final evidence set. GPT-5.4 with w/o semantic gate obtains 70.86\% exact / 86.29\% binary accuracy, and Qwen3.5-9B obtains 70.86\% exact / 86.00\% binary accuracy. These drops show that a Lean proof or checker success is useful only when the generated formal statement remains aligned with the original natural-language obligation. Without the semantic gate, formally valid but semantically shifted statements can contaminate the final judgment.

\ifdefined\arxivpublic
  \subsection{Label Balance and End-to-End Error Decomposition}
\label{app:error-decomposition}

\textsc{ProofLoc-Olympiad} contains 153 correct and 197 incorrect proofs;
\textsc{ProofLoc-University} contains 130 correct and 70 incorrect proofs.
Table~\ref{tab:end-to-end-confusion} reports the complete GPT-5.4 confusion
matrices.  We treat an incorrect proof as the positive class: FP is a correct
proof flagged as incorrect, and FN is an incorrect proof accepted as correct.

\begin{table}[h]
\centering
\caption{Complete GPT-5.4 correct/incorrect confusion matrices.}
\label{tab:end-to-end-confusion}
\footnotesize
\begin{tabular}{@{}llrrr@{}}
\toprule
Dataset & Method & TP/TN/FP/FN & FPR & FNR \\
\midrule
Olympiad-350 & Direct@1 & 189/120/33/8 & 21.57\% & 4.06\% \\
Olympiad-350 & \textsc{FaithSieve} & 189/139/14/8 & 9.15\% & 4.06\% \\
University-200 & Direct@1 & 66/108/22/4 & 16.92\% & 5.71\% \\
University-200 & \textsc{FaithSieve} & 64/121/9/6 & 6.92\% & 8.57\% \\
\bottomrule
\end{tabular}
\end{table}

\textsc{FaithSieve} reduces the false-positive rate from 21.57\% to 9.15\%
on Olympiad and from 16.92\% to 6.92\% on University.  Table~\ref{tab:exact-error-types}
separates binary errors from cases where the binary verdict is correct but the
first-error position is mislocalized.

\begin{table}[h]
\centering
\caption{Composition of the remaining GPT-5.4 \textsc{FaithSieve} Exact errors.}
\label{tab:exact-error-types}
\footnotesize
\begin{tabular}{@{}lrrrr@{}}
\toprule
Dataset & Exact errors & FP & FN & Binary correct, location wrong \\
\midrule
Olympiad-350 & 65 & 14 (21.5\%) & 8 (12.3\%) & 43 (66.2\%) \\
University-200 & 31 & 9 (29.0\%) & 6 (19.4\%) & 16 (51.6\%) \\
\bottomrule
\end{tabular}
\end{table}

The largest residual category on both datasets is therefore localization rather
than binary proof classification.  On University-200, the paired Exact table
contains 139 cases where both Direct@1 and \textsc{FaithSieve} are correct, 30
where only \textsc{FaithSieve} is correct, 11 where only Direct@1 is correct,
and 20 where both are wrong.  This separates the 9.5-point Exact gain over
Direct@1 from the remaining 15.5-point gap to perfect localization.

\subsection{Multi-Sample, Deliberative, and External Baselines}
\label{app:additional-baselines}

We added multi-sample direct judging and a deliberative baseline under the same
first-error localization interface.  Direct@5 aggregates five independent
judgments.  The deliberative baseline uses two independent critics followed by
one adjudicator.  Within each dataset, all methods use the same problems and
backbone: GPT-5.4 on the complete University-200 set and Qwen3.5-9B on the
tested 200-example BIG-Bench Mistake subset.

\begin{table}[h]
\centering
\caption{Additional natural-language baselines.  The BIG-Bench Mistake result uses the tested 200-example subset.}
\label{tab:additional-nl-baselines}
\footnotesize
\begin{tabular}{@{}llrrr@{}}
\toprule
Dataset / backbone & Method & $N$ & Exact (\%) & Binary (\%) \\
\midrule
University / GPT-5.4 & \textsc{FaithSieve} & 200 & \textbf{84.5} & \textbf{92.5} \\
University / GPT-5.4 & Direct@1 & 200 & 75.0 & 87.0 \\
University / GPT-5.4 & Direct@5 & 200 & 76.0 & 85.5 \\
University / GPT-5.4 & Two critics + adjudicator & 200 & 76.0 & 86.5 \\
\midrule
BIG-Bench / Qwen3.5-9B & \textsc{FaithSieve} & 200 & \textbf{61.0} & \textbf{74.5} \\
BIG-Bench / Qwen3.5-9B & Direct@1 & 200 & 51.0 & 56.5 \\
BIG-Bench / Qwen3.5-9B & Direct@5 & 200 & 51.5 & 55.0 \\
BIG-Bench / Qwen3.5-9B & Two critics + adjudicator & 200 & 54.5 & 64.0 \\
\bottomrule
\end{tabular}
\end{table}

On University Exact, \textsc{FaithSieve} improves over both Direct@5 and the
critic baseline by 8.5 points.  The paired 95\% CI for both comparisons is
$[+2.5,+14.5]$ and McNemar's $p=0.00948$.  The corresponding BIG-Bench Exact
gains are 9.5 and 6.5 points.

We also evaluated three public process reward models as \emph{off-the-shelf
transfer diagnostics}.  Each PRM scores the original proof steps, predicts the
first step below its frozen threshold, and predicts \texttt{correct} if every
step passes.  Thresholds were calibrated externally on ProcessBench
GSM8K-400 and frozen before evaluation: 0.904296875 for
Qwen2.5-Math-PRM-7B, 0.34375 for Skywork-o1-Open-PRM, and
0.5007799117379625 for EurusPRM-Stage2.

\begin{table}[h]
\centering
\caption{Off-the-shelf PRM transfer under the same first-error interface.}
\label{tab:prm-transfer}
\scriptsize
\resizebox{\linewidth}{!}{%
\begin{tabular}{@{}lcrrrr@{}}
\toprule
Method & Model size & University Exact & University Binary & BIG-Bench Exact & BIG-Bench Binary \\
\midrule
\textsc{FaithSieve} / Qwen3.5-9B & 9B & \textbf{69.5} & \textbf{78.5} & \textbf{61.0} & \textbf{74.5} \\
Qwen3.5-9B Direct & 9B & 61.0 & 76.5 & 51.0 & 56.5 \\
Qwen2.5-Math-PRM-7B & 7B & 47.5 & 64.0 & 56.5 & 74.0 \\
Skywork-o1-Open-PRM & 7B & 47.5 & 64.5 & 14.0 & 61.0 \\
EurusPRM-Stage2 & 7B policy + 7B reference & 12.0 & 37.0 & 9.5 & 51.5 \\
\bottomrule
\end{tabular}%
}
\end{table}

This table is not a model-capability- or token-budget-matched comparison.  It
tests whether existing public mathematical PRMs transfer directly to a fixed
proof's first-error localization task.  Standard Best-of-$N$ instead ranks
$N$ candidate solutions.  Here Direct@$N$ produces multiple judgments of one
fixed proof; a solution-step PRM applied to that same proof produces the same
score vector for every Direct sample.  Applying such a PRM to Direct's
evaluator rationales would instead be an uncalibrated use on evaluator traces,
which these PRMs were not trained to score.

The BIG-Bench Mistake subset contains 200 examples: 80 Multistep Arithmetic,
60 Tracking Shuffled Objects, and 60 Dyck Languages examples, with 100 correct
and 100 incorrect proofs.  It was sampled because a complete end-to-end run
over all 2,186 examples was not feasible within the available compute budget.

\subsection{Token Cost and High-Budget Controls}
\label{app:cost-controls}

\paragraph{Recorded pipeline cost.}
FaithSieve has high model-token consumption because it includes agentic
formalization and proof generation.  Under a fixed 50-problem cost-evaluation
setting, total tokens were the unweighted sum of recorded input and output
tokens.

\begin{table}[h]
\centering
\caption{Average recorded token consumption in the fixed 50-problem cost evaluation.}
\label{tab:recorded-cost}
\footnotesize
\begin{tabular}{@{}lrrr@{}}
\toprule
Model & Direct@1 tokens/problem & \textsc{FaithSieve} tokens/problem & Ratio \\
\midrule
GPT-5.4 & 910.74 & 124,970.56 & 137.22$\times$ \\
Qwen3.5-9B & 1,078.34 & 164,025.20 & 152.11$\times$ \\
\bottomrule
\end{tabular}
\end{table}

\paragraph{Same-backbone high-budget controls.}
We added Direct Majority scaling and two high-budget NL-only agents on all 200
examples of the same pre-fixed BIG-Bench Mistake subset.  Every method in
Table~\ref{tab:token-budget-controls} uses the same Qwen3.5-9B checkpoint with
thinking disabled.  The controls use a 300-second per-request timeout; ground
truth is joined only after outputs are frozen.  Direct Majority uses
temperature 0.6 and a 64-token output limit.  The two structured NL-only
agents use temperature 0.1 and stage-specific output limits of 64--3072.
Other decoding settings follow service defaults.

\begin{table}[h]
\centering
\caption{Same-data, same-backbone token-budget controls on BIG-Bench-200.}
\label{tab:token-budget-controls}
\footnotesize
\begin{tabular}{@{}lrrr@{}}
\toprule
Method & Exact (\%) & Binary (\%) & FaithSieve token share \\
\midrule
\textsc{FaithSieve} / Qwen3.5-9B & \textbf{61.0} & \textbf{74.5} & 100.00\% \\
Direct Majority@1 (pool mean) & 51.6 & 55.7 & 0.42\% \\
Direct Majority@3 & 52.2 & 55.5 & 1.25\% \\
Direct Majority@5 & 52.0 & 55.1 & 2.08\% \\
Direct Majority@9 & 52.0 & 54.6 & 3.75\% \\
Direct Majority@17 & 51.8 & 54.1 & 7.08\% \\
Direct Majority@33 & 51.6 & 53.8 & 13.74\% \\
Direct Majority@65 & 51.5 & 53.6 & 27.07\% \\
Direct Majority@129 & 51.4 & 53.4 & 53.73\% \\
Direct Majority@241 & 51.0 & 53.0 & 100.37\% \\
Context-Isolated Reviewer Panel & 52.0 & 56.0 & 124.41\% \\
NL-Only Decomposition + Challenge-Defense & 55.5 & 57.0 & 82.85\% \\
\bottomrule
\end{tabular}
\end{table}

Direct Majority@$N$ aggregates normalized labels in
$\{\texttt{correct},\texttt{step }k\}$ by plurality.  Ties select the earliest
numeric step; \texttt{correct} is selected only when it is the unique top
label.  For $N<241$, entries are means over 1,000 fixed-seed samples without
replacement from a frozen 241-vote pool; $N=241$ uses the complete pool.
The slight decline with larger $N$ reflects systematic modal bias rather than
an aggregation anomaly: 90/200 problems receive the same label in all 241
samples, another 44 contain the gold label but not as the modal label, and
there are no modal ties at $N=65,129,$ or 241.

The Context-Isolated Reviewer Panel creates three independent roles: validity,
localization, and counterexample review.  Each role contains 32 separately
sampled reviewers with isolated contexts, followed by role summaries, two
adjudication rounds, final synthesis, consistency checking, and label
normalization.  The NL-Only Decomposition + Challenge-Defense agent decomposes
the proof while retaining source-step mappings, selects at most four candidate
first errors, runs challenger/defender/judge analysis plus 12 independent
audits per candidate, and then performs global consistency, critique, revision,
and final synthesis.  Neither agent uses Lean, a PRM, or FaithSieve artifacts.

These controls do not make FaithSieve inexpensive.  Instead, they show that on
the same data and backbone, reallocating a comparable token budget to repeated
natural-language judgments or structured NL-only agents did not reproduce the
gain from adding locally checkable Lean evidence.

\subsection{Domain-Level Uncertainty}
\label{app:domain-uncertainty}

We report the repeated domain-level analysis on University-200.  Table~\ref{tab:domain-exact-ci}
gives sample sizes and two-sided Wilson 95\% confidence intervals for Exact
accuracy.  All methods use GPT-5.4; the FaithSieve point estimates are the
original University-200 results.

\begin{table}[h]
\centering
\caption{University-200 Exact accuracy with two-sided Wilson 95\% confidence intervals.}
\label{tab:domain-exact-ci}
\scriptsize
\resizebox{\linewidth}{!}{%
\begin{tabular}{@{}lrrrr@{}}
\toprule
Domain & $n$ & FaithSieve & Direct@5 & Two critics + adjudicator \\
\midrule
Topology & 20 & 80.0 [58.4, 91.9] & 65.0 [43.3, 81.9] & 65.0 [43.3, 81.9] \\
Linear Algebra & 40 & 100.0 [91.2, 100.0] & 92.5 [80.1, 97.4] & 95.0 [83.5, 98.6] \\
Abstract Algebra & 40 & 97.5 [87.1, 99.6] & 90.0 [76.9, 96.0] & 87.5 [73.9, 94.5] \\
Convex Analysis & 30 & 60.0 [42.3, 75.4] & 50.0 [33.2, 66.8] & 56.7 [39.2, 72.6] \\
Convex Optimization & 20 & 80.0 [58.4, 91.9] & 70.0 [48.1, 85.5] & 55.0 [34.2, 74.2] \\
Real Analysis & 50 & 80.0 [67.0, 88.8] & 74.0 [60.4, 84.1] & 76.0 [62.6, 85.7] \\
\bottomrule
\end{tabular}%
}
\end{table}

FaithSieve is relatively high on Linear Algebra and Abstract Algebra.  A
plausible explanation is that these domains often expose clearer algebraic
objects, theorem dependencies, and local derivation structures, which align
well with fine-grained decomposition, statement formalization, and Lean
verification.  The intervals remain important because several domains contain
only 20--40 examples.

\subsection{Decoding Configuration of the Original University Experiment}
\label{app:may-reproducibility}

Table~\ref{tab:may-decoding} reports the frozen stage-specific settings of the
original University-200 experiment.  It does not mix in configurations from
additional baselines.  Output limits are reported only where they were
explicitly set at the call site; otherwise the provider default was used.

\begin{table}[h]
\centering
\caption{Decoding configuration of the original University-200 experiment.}
\label{tab:may-decoding}
\footnotesize
\begin{tabular}{@{}lrr@{}}
\toprule
Method / stage & Temperature & Max output tokens \\
\midrule
Direct@1 & 0 & 64 \\
Decomposition generation & 0.3 & default \\
Decomposition verification / deduplication & 0.2 & default \\
Informal-tree generation & 0.7 & default \\
Informal-tree verification & 0.2 & default \\
Suspicion Search & 0.1 & 1600 \\
Local Review & 0.1 & 1600 \\
Primary Final Synthesis & 0.1 & 1600 \\
Final consistency correction / JSON repair & 0 & 700 \\
Strict-JSON retry & 0 & 1600 \\
\bottomrule
\end{tabular}
\end{table}

The shared settings were: \texttt{top\_p}, request seed, frequency penalty,
presence penalty, and stop all used provider defaults.  These settings were
fixed throughout the complete University-200 evaluation.

\fi

\section{Limitations and Broader Impacts}
\label{app:limitations-broader-impacts}

\subsection{Limitations}
\label{app:limitations}

Although our method improves exact and binary accuracy for first-error localization on \textsc{ProofLoc-Olympiad}, it is limited by the cost and coverage of formal validation. Lean-assisted verification involves statement generation, compile repair, proof search, artifact checking, and evidence normalization; any stage may fail or time out. The method also depends on current LLM agents for autoformalization and proof repair, so unfamiliar Lean representations or mismatches between mathlib definitions and natural-language conventions can turn an otherwise checkable claim into inconclusive evidence.

A second limitation is domain coverage. \textsc{ProofLoc-Olympiad} focuses on algebra and number theory, where local transformations often form clear typed obligations. Geometry and combinatorics require diagrammatic objects, construction relations, counting models, and problem-specific encodings, so failures may reflect modeling choices or library coverage. \textsc{ProofLoc-University} expands the evaluation to six abstract domains, but broader university-level mathematics still requires additional annotation and verification.

\subsection{Broader Impacts}
\label{app:broader-impacts}

This work aims to move mathematical reasoning systems from producing plausible final answers toward providing local evidence that is traceable, checkable, and interpretable. Such evidence may support more reliable AI proof assistants, step-level feedback in mathematical education, quality control for generated proofs, and better interfaces between informal mathematical writing and formalized mathematics.

At the same time, \textsc{FaithSieve} should not be viewed as an automatic replacement for human mathematical judgment. It is better understood as an auditing aid: it helps identify high-risk proof steps, filters unfaithful formalizations, and organizes checkable evidence in the context of the original proof. In educational settings, this distinction is especially important: local feedback should help students inspect and repair reasoning, rather than simply assign authoritative correctness labels.

The main risk is over-reliance on incomplete formal evidence. A successful Lean proof is useful only when the generated statement faithfully represents the original obligation, and an inconclusive result does not imply that the proof step is wrong. For this reason, the system explicitly separates semantic faithfulness, formal validation status, and final benchmark-level prediction. We expect responsible uses to preserve this evidence trail and to present uncertain cases as uncertain, especially in high-stakes mathematical review or instruction.

\section{Case Studies: Natural-Language Context for Representative Lean Statements}
\label{app:case-studies}

The following examples expose the source natural-language claim, the extracted
local obligation, and an ASCII-rendered Lean target.  The execution records
retain the complete problem, full natural-language proof, source-step mapping,
local context, generated statement, and proof result.  All six proof scripts
were rechecked successfully with Lean.  The first two targets prove explicit
negations of incorrect source claims; the remaining four prove correct local
obligations.  A successful local theorem is evidence only for its corresponding
NL--FL pair, not a certificate for the entire proof.

\paragraph{Example 1: polynomial witnesses (caught error).}
The problem asks for the smallest number of rational-coefficient polynomial
squares needed to represent $X^2+7$.  The source step claims that
$f_1=X^2+2$ and $f_2=f_3=f_4=-1$ satisfy the identity, thereby making four
summands feasible.  The local validation target retains all four concrete
witnesses and proves that no such instantiation satisfies the asserted
identity.

\begin{PromptBox}
theorem polynomial_witness_neg :
  Not (Exists fun f1 : Polynomial Rat =>
    Exists fun f2 : Polynomial Rat =>
    Exists fun f3 : Polynomial Rat =>
    Exists fun f4 : Polynomial Rat =>
      Polynomial.X ^ 2 + 7 = f1 ^ 2 + f2 ^ 2 + f3 ^ 2 + f4 ^ 2 /\
      f1 = Polynomial.X ^ 2 + 2 /\
      f2 = -1 /\ f3 = -1 /\ f4 = -1)
\end{PromptBox}

At $X=1$, the asserted identity gives 8 on the left and 12 on the right;
equivalently, the polynomial coefficient comparison fails because
$(X^2+2)^2$ introduces an $X^4$ term.  Lean verifies the negation and the
result is mapped back to the source claim.

\paragraph{Example 2: quadratic coefficient (caught error).}
The local context contains $T=8$, $c/a=T$, the point-value equations
$f(-2)=20$ and $f(1)=14$, and the simplified equations
$12a-2b=20$ and $9a+b=14$.  The source step concludes $a=4/3$; the extracted
obligation asks whether that conclusion follows.  The validation branch proves
its negation while retaining the inherited equations.

\begin{PromptBox}
theorem quadratic_coefficient_neg
    (a b c T : Real) (f : Real -> Real)
    (hT : T = 8) (hf : forall x, f x = a * x ^ 2 + b * x + c)
    (hprod : c / a = T) (hpt1 : f (-2) = 20) (hpt2 : f 1 = 14)
    (heq1 : 4 * a - 2 * b + c = 20)
    (heq2 : a + b + c = 14)
    (hcrel : c = 8 * a)
    (hsimp1 : 12 * a - 2 * b = 20)
    (hsimp2 : 9 * a + b = 14) :
    Not (a = 4 / 3)
\end{PromptBox}

Adding twice the second simplified equation to the first yields $30a=24$, so
the consistent value is $a=4/5$, not $4/3$.

\paragraph{Example 3: functional equation (proved local step).}
The problem gives $f(2x+3)=2f(x)+3$ and $f(0)=6$.  After deriving $f(3)=15$,
the source step substitutes $x=3$ and concludes $f(9)=33$.  The theorem keeps
the recurrence, initial value, and previously derived local fact.

\begin{PromptBox}
theorem functional_step
    (f : Real -> Real)
    (hprop : forall x, f (2 * x + 3) = 2 * f x + 3)
    (hbase : f 0 = 6) (hf3 : f 3 = 15) :
    f 9 = 33
\end{PromptBox}

\paragraph{Example 4: logarithm change of base (proved local step).}
The source proof rewrites $\log_4 x$ as
$\tfrac12\log_2 x$ while solving a three-equation logarithmic system.  The
local statement retains positivity and the original system rather than
checking an isolated identity without its proof context.

\begin{PromptBox}
theorem log_change_of_base
    (x y z : Real) (hx : x > 0) (hy : y > 0) (hz : z > 0)
    (hsys : Real.logb 4 x + Real.logb 8 (y * z) = 2 /\
      Real.logb 4 y + Real.logb 8 (x * z) = 4 /\
      Real.logb 4 z + Real.logb 8 (x * y) = 5) :
    Real.logb 4 x = (1 / 2) * Real.logb 2 x
\end{PromptBox}

\paragraph{Example 5: real-exponentiation rewrite (proved local step).}
The source equation is
$2^{x+2}5^{6-x}=10^{x^2}$.  The local obligation checks the rewrite of the
right-hand side into $2^{x^2}5^{x^2}$ while retaining the original equation as
an assumption.

\begin{PromptBox}
theorem exponent_rewrite (x : Real)
    (hgiven : (2 : Real) ^ (x + 2) * (5 : Real) ^ (6 - x) =
      (10 : Real) ^ (x ^ 2)) :
    (2 : Real) ^ (x + 2) * (5 : Real) ^ (6 - x) =
      (2 : Real) ^ (x ^ 2) * (5 : Real) ^ (x ^ 2)
\end{PromptBox}

\paragraph{Example 6: integer arithmetic (proved local value).}
In BIG-Bench Mistake example 71, the source proof computes
$B=0\cdot5-(-1)-(-5)=6$.  The extracted statement has the same concrete
arithmetic target and is discharged by normalization.

\begin{PromptBox}
theorem integer_arithmetic :
  (0 : Int) * 5 - (-1) - (-5) = 6
\end{PromptBox}

\end{document}